\documentclass[nohyperref]{article}

\usepackage{microtype}
\usepackage{graphicx}
\usepackage{float}
\usepackage{subfigure}
\usepackage{booktabs} 
\usepackage{amsmath}
\usepackage{amssymb}
\usepackage{mathtools}
\usepackage{amsthm}
\usepackage[tableposition=top]{caption}
\usepackage{float}
\floatstyle{plaintop}
\restylefloat{table}
\usepackage{hyperref}
\usepackage[capitalize,noabbrev]{cleveref}

\usepackage[accepted]{icml2022} 

\newcommand{\QED}{\hfill\ensuremath{\square}}

\theoremstyle{plain}
\newtheorem{theorem}{Theorem}
\newtheorem{proposition}[theorem]{Proposition}

\theoremstyle{definition}
\newtheorem{definition}{Definition}

\theoremstyle{remark}

\usepackage{amsmath}

\icmltitlerunning{Utilizing Expert Features for Contrastive Learning of Time-Series Representations}

\usepackage{cleveref}
\Crefname{equation}{Eq.}{Eqs.}
\Crefname{figure}{Fig.}{Figs.}
\Crefname{tabular}{Tab.}{Tabs.}

\begin{document}

\twocolumn[
\icmltitle{Utilizing Expert Features for Contrastive Learning\\of Time-Series Representations}


\icmlsetsymbol{equal}{*}

\begin{icmlauthorlist}
\icmlauthor{Manuel Nonnenmacher}{equal,bcai,us}
\icmlauthor{Lukas Oldenburg}{bcai}
\icmlauthor{Ingo Steinwart}{us}
\icmlauthor{David Reeb}{bcai}
\end{icmlauthorlist}

\icmlaffiliation{bcai}{Bosch Center for Artificial Intelligence (BCAI), Robert Bosch GmbH, Renningen, Germany}
\icmlaffiliation{us}{Institute for Stochastics and Applications, University of Stuttgart, Stuttgart, Germany}

\icmlcorrespondingauthor{Manuel Nonnenmacher}{manuel.nonnenmacher@de.bosch.com}

\icmlkeywords{Time-Series ,Representation Learning, Embeddings, Contrastive Learning, Expert Features}

\vskip 0.3in ]



\printAffiliationsAndNotice{}  

\begin{abstract}
We present an approach that incorporates expert knowledge for time-series representation learning. Our method employs expert features to replace the commonly used data transformations in previous contrastive learning approaches. We do this since time-series data frequently stems from the industrial or medical field where expert features are often available from domain experts, while transformations are generally elusive for time-series data. We start by proposing two properties that useful time-series representations should fulfill and show that current representation learning approaches do not ensure these properties. We therefore devise ExpCLR, a novel contrastive learning approach built on an objective that utilizes expert features to encourage both properties for the learned representation. Finally, we demonstrate on three real-world time-series datasets that {ExpCLR} surpasses several state-of-the-art methods for both unsupervised and semi-supervised representation learning.



\end{abstract}

\section{Introduction}\label{sec:intro}
Contrastive learning (CL) has led to significant advances in the field of representation learning \citep{bromley1993signature,hadsell2006dimensionality}. Recently, the introduction of the NCE-loss \citep{gutmann2010noise} and its variants \citep{oord2018representation,chen2020simple,he2020momentum,grill2020bootstrap,zbontar2021barlow} pushed the sate-of-the-art for representation learning forward. These loss functions make use of different views of the data, which are transformations of the data that leave the individual label of each sample invariant. During training, by minimizing the loss function, the different views of the same sample are then pushed towards the same representations. The aim is often to learn representations where samples with the same label are close and samples with different labels are well separated. 

Such an approach is particularly successful in the vision domain, where different views (or transformations) of the data are readily available. Contrary to that, in other data domains such as time-series data, these transformations are not as easily found or requiring deep task and domain knowledge. Further, purely relying on the different views of the data does not generally lead to good representations of the data for time-series \cite{iwana2021empirical,eldele2021time}. On the other hand, time-series datasets often stem from an industrial or medical domain, where domain experts can usually provide expert features. Expert features are therefore often readily available while access to good data transformations is generally not.

Therefore, the goal of this work is to make use of these expert features in order to find good embeddings or representations of time-series datasets utilizing expert features instead of data transformations. The use of expert features introduces several new challenges such as, which loss-function one should use and how to define good expert features. Apart from these challenges the use of expert features also circumvents several problems present in transformation based CL approaches such as the sampling bias for negative samples and the challenge of finding good transformations of the data, which can be critical for CL \citep{tian2020makes, tamkin2020viewmaker}.

While expert features can usually be provided by domain experts, it remains open to find a new loss function to train using these expert features. One major challenge is that these expert features are usually not discrete class labels but continuous features, such as temperature, speed or energy values. These expert features could stem from additional measurements, (potentially expensive) simulations, or be calculated from expert mappings that use domain knowledge, directly. Therefore, using existing loss functions such as the NCE-loss or the pair-loss is not straightforwardly possible. To guide the design of a loss function, we will first establish two properties that a representations based on expert features should fulfill to be useful for a range of downstream tasks. First, we want the representations of two input samples with similar expert features to be close to each other; and second, we want a pair of samples with very different expert features to be far apart in the representation space. These two principles will help us design a loss function that encourages a representation with desirable properties.

While a representation or embedding can be used for a variety of downstream tasks (i.e. predictive models, outlier detection, active learning, identifying similar samples), the two properties defined earlier assure that the representation is suited for all of these tasks. On the other hand, good performance on one of these tasks does not assure that the two properties are fulfilled (see Sec.\ \ref{sec:fail_feed_forward}) and therefore does also not give any performance guarantees for any of the other tasks.

Our contributions are as follows:
\begin{itemize}
    \item  We introduce two properties that a useful data representation based on expert features should satisfy, and show that only excelling in a single downstream task does not necessarily lead to a representation having the desired properties.

    \item Next, we introduce a novel loss function which is able to utilize continuous expert features to learn a useful representation of the data. We show that the representation obtained by minimizing our loss function attains both previously defined properties. We name the method utilizing this novel loss function ExpCLR. 
    
    \item Finally, we compare our approach to several state-of-the-art approaches in the unsupervised and semi-supervised setting. For each of the three real-world datasets, our method outperforms or is on par with all other methods for several evaluation metrics.
\end{itemize}

PyTorch code implementing our method is provided at \href{https://github.com/boschresearch/expclr}{https://github.com/boschresearch/expclr}.

\section{Related Work}\label{sec:related}
Our work draws on existing literature including self-supervised representation learning, knowledge distillation, metric learning and works that try to combine expert knowledge with neural networks. 

While our loss is inspired by the pair-loss function \cite{hadsell2006dimensionality,le2020contrastive} it makes use of components used by many other contrastive loss functions \citep{gutmann2010noise,chen2020simple,zbontar2021barlow,grill2020bootstrap,he2020momentum}. Although having a similar objective to ours \cite{wang2020understanding}, these loss functions use augmentations to create positive label pairs, while we make use of expert features to determine the similarity of pairs allowing the use of continuous labels or features. Further, some of these works suffer from the negative sample bias \cite{chuang2020debiased}, a problem which our loss function does not encounter.
Additionally, the loss function proposed by us can also be used with discrete labels in a setting similar to supervised CL \cite{khosla2020supervised}.

Most of the CL approaches mentioned in the previous section are applied in the vision domain, where most of the progress was made. Recently, an increasing number of works apply CL to time-series data. An early work was Contrastive Predictive Coding (CPC) \cite{oord2018representation}. More recent works try to combine classical CL approaches with time-series specific training objectives and augmentations such as slicing \citep{tonekaboni2020unsupervised,franceschi2019unsupervised,zheng2021weakly}, forecasting \citep{eldele2021time} and neural processes \citep{kallidromitis2021contrastive}.

Two fields that are also closely linked to our approach are Deep Metric Learning and Knowledge Distillation, where a small student network tries to imitate the output or metric induced by a larger teacher network \cite{park2019relational,kim2020proxy}. While traditional deep metric learning does not make use of the inherent continuous nature of the teacher model, recent works have tried go beyond binary supervision and make use of this \cite{kim2019deep,kim2021embedding}. Also closely linked to this field and our work is the field of Knowledge Distillation \cite{gou2021knowledge}, especially the works of \citet{park2019relational} and \citet{yu2019learning}, which also take geometric relations of the teacher model into account. While these works use similar loss functions, which can also make use of the continuous nature of the expert features, our loss function leads to superior performance (Sec.\ \ref{sec:experiments}). Further, we have a different goal, which is to precondition our representations by using the expert features to obtain a representation with favorable properties, which can then be used for a multitude of downstream tasks.

Lastly our method tries to incorporate expert knowledge with neural network training. There are several other works aiming to achieve this  \cite{chattha2019kinn,hu2016harnessing}. The two most relevant works are SleepPriorCL \citep{zhang2021sleeppriorcl} and TREBA \citep{sun2021task}. Similar to our work, TREBA and SleepPriorCL try to learn an embedding for trajectories to improve labeling-efficiency. In contrast to our work, TREBA not only proposes a contrastive loss to
handle continuous expert features but further combines contrastive learning with several other training objectives such as reconstruction and consistency. While both SleepPriorCL and TREBA aim to use expert features to create pseudo-labels for unsupervised and semi-supervised representation learning by discretizing the continuous expert features, ExpCLR is able to assess expert feature distances continuously into its objective; this avoids information loss coming from the binary positive vs.\ negative grouping.

\section{Method}\label{sec:methods}
\subsection{Contrastive Learning Setting}\label{sec:setting}
A neural network encoder (or simply \emph{encoder}) maps samples from the input domain $x\in{\mathbb R}^{c \times T}$, where $c$ denotes the number of input channels and $T$ the number of time steps of each sample\footnote{For notational simplicity we consider here time-series of fixed length, while our formalism is straightforward to extend to varying-length time-series.}, to an embedding or \emph{representation} $E_\theta(x)\in{\mathbb R}^e$. The encoder's parameters (weights) are $\theta\in{\mathbb R}^P$, which are updated in supervised learning by minimizing a loss-function ${\mathcal L}(E_\theta(X),Y)$ on the training set $(X,Y)=\{(x_1,y_1),\dots,(x_N,y_N)\}$, where the labels $y_i$ are often discrete classes $y \in \{1, \dots, C\}$. 

In \emph{contrastive representation learning} the loss function is chosen in such a way that representations of samples with the same class label are pulled closer to each other w.r.t.\ the Euclidean norm, while samples with different class labels are pushed away from each other. This can be achieved by minimizing a contrastive loss function, e.g.\ the ``triplet-Loss'' \cite{chechik2010large} or the ``NCE-Loss'' \cite{gutmann2010noise}. Another prominent contrastive loss is the so called \emph{pair-loss} function \cite{hadsell2006dimensionality,le2020contrastive}:
\begin{align}\label{eq:pair_loss}
    &{\mathcal L}_{pair}(E(X),Y)  \\ &~~=\frac{1}{N^2} \sum_{i,j=1}^N s_{ij}D_{ij}^2
    +\max\left\{0,(1-s_{ij})^2\Delta^2-D_{ij}^2\right\},\nonumber
\end{align}
where $s_{ij}$ is the discrete similarity measure defined by $s_{ij}:= \delta_{y_i=y_j}$, $\Delta \in \mathbb{R_+}$ is a hyperparameter, and $D_{ij}:= \|E(x_i)-E(x_j)\|_2$ denotes the Euclidean distance. While the first term in the sum is responsible for pulling closer together the representations of similar points, i.e.\ pairs with the same labels, the second term aims to push representations of dissimilar pairs to a distance of at least $\Delta$. So far, this is the \emph{supervised setting} of CL, where all labels are provided. Next, we describe the \emph{semi-supervised setting}, where labels are provided only for a fraction of the dataset, and the \emph{unsupervised} setting as the extreme case with no labels available.

In the unsupervised setting, most CL algorithms make use of transformations $\{m_1,\dots, m_V\}$ with  $m_v: {\mathbb R}^{c \times T} \to {\mathbb R}^{c \times T}$, which leave the class label invariant. These transformations can then be used to create so-called ``views'' $x_i^v:=m_v(x^0_i)$ of a data sample $x^0_i:=x_i$. By assumption, all $x_i^v$ have the same $y$-label as the original $x^0_i$ (even if this label is unknown) and therefore discrete similarity measure $s_{i,i}^{0,v}=1$. A number of other randomly selected data samples $x^0_j\ (j\neq i)$, e.g.\ the other samples in the batch, are then considered negative samples with similarity $s_{i,j}^{0,0}=0$, such that one can write down a CL loss like (Eq. \ref{eq:pair_loss}) using these transformations. The expert knowledge used for unsupervised learning is thus the transformations $m_v$ leaving class labels invariant. The training pushes the encoder towards being invariant w.r.t.\ the transformations $m_v$. In the vision domain, a great number of sensible transformations $m_v$ are known (cropping, rotation, translation, etc.).

In the time-series domain, however, finding invariant transformations is less intuitive, and one can easily be led astray \cite{iwana2021empirical}. But since many time-series datasets come from fields like industry or medicine in particular, expert features are often readily available from domain experts. These expert features may be discrete or (more usually) continuous, and could e.g.\ be calculated from the input time-series by an expert mapping $f:{\mathbb R}^{c \times T} \to {\mathbb R}^{d}$, i.e.\ $f_i:= f(x_i)$, and/or be available from additional measurement sensors on the training data. In our work we assume to be given a set of expert features\footnote{Note, we do not assume the expert mapping $f$ to be given. It is unavailable e.g.\ in the HAR dataset (Sec.\ \ref{sec:experiments}).} $F=\{f_1,\dots,f_N\}$ for the training inputs $X$. The full dataset to train our embedding is thus $(X,F,Y)$, where $Y$ may contain label information for any percentage of input data points, ranging between the supervised and the fully unsupervised setting.

\subsection{Desiderata for Representations}\label{sec:embedding_properties}
We aim to employ the given expert features $F$ in a way to learn a good representation of the input time-series. To see how to best utilize $F$, we first discuss what properties a useful representation should have  w.r.t.\ the expert features. We follow the general ideal behind CL, which is to push points with the same (resp.\ different) labels together (resp.\ apart). For continuous-valued expert features $f_i\in \mathbb{R}^d$, this motivates an encoder $E$ with the following properties:
\begin{itemize}
    \item[\textbf{(P1)}]If expert features of two points are similar, i.e.\ $\|f_i-f_j\|_2$ is small, then $\|E(x_i)-E(x_j)\|_2$ should also be small, i.e.\ the corresponding representations be similar.
    \item[\textbf{(P2)}] If  $\|f_i-f_j\|_2$ is large, then $\|E(x_i)-E(x_j)\|_2$ should also be large.
\end{itemize}
Here, $x_i,x_j$ are any two samples from $X$, and $f_i,f_j$ their given expert features. While both properties can be important for predictive models, (P1) is especially important for outlier detection, while (P2) is important for identifying similar samples and (safe) active learning.

The question may arise why one cannot directly use the expert features $f$ as the representation $E(x)$, which would satisfy both properties trivially. To start, note that we do \emph{not} assume the full expert mapping $f$ to be available, but merely the features $f_i$ for the given inputs $X$; i.e.\ one couldn't evaluate $E(x_{test})$ at test inputs with such a prescription. And even if $f$ were available, downstream tasks often benefit from fine-tuning the encoding function $E=E_\theta$ further (Sec.\ \ref{exp:semi-supervised}), which is generally not possible or successful with the expert feature mapping $f$. Second, the mapping $f$ would not allow to freely choose the dimension $e$ for the representation space, but bind it to the feature dimension $d$. Finally, we observe in experiments that a learned $E$ allows to exceed the performance over the original features $f$ in downstream tasks even for the unsupervised setting (Tab.\ \ref{tab:unsupervised_results}).

We formalize the properties (P1) and (P2) by defining \emph{bilipschitz representations} w.r.t.\ the given set of expert features:
\begin{definition}[bilipschitz representation]\label{definition1}A representation $E:{\mathbb R}^{c \times T} \to {\mathbb R}^{e}$ is called a $[l_-,l_+]$-bilipschitz representation for $0<l_-\leq l_+<\infty$ if $\forall i,j \in \{1,\dots,N\}$:
\begin{align*}
    l_- \| E_i-E_j\|_2 \leq \|f_i-f_j\|_2 \leq l_+ \| E_i-E_j\|_2,
\end{align*}
where $E_i := E(x_i)$ and $E_j := E(x_j)$.
\end{definition}
Note that we require (and are able to evaluate) the condition in Def.\ \ref{definition1} only on the training set $X$ and \emph{not} on all potential input points $x\in{\mathbb R}^{c\times T}$. But from this, one can derive statistical bounds on the pair-Lipschitz constant $\|f-f'\|_2/ \| E(x)-E(x')\|_2$ for test points $x,x'$ (App.\ \ref{sec:statistical_bounds}), even though it is generally impossible to satisfy Def.\ \ref{definition1} on an infinite set of inputs when the feature dimension $d>e$ exceeds the representation dimension.

The larger $l_-$ and the smaller $l_+$ is in Def.\ \ref{definition1}, the better are the guarantees one can provide for (P1) and (P2), respectively. In the ideal case we have $l_-=l_+$, i.e.\ the Euclidean distance in the representation space is proportional to the distance of the expert features.

\subsection{Learning Representations via Feedforward Models}\label{sec:fail_feed_forward}
Having established the properties a useful representation should have, we may ask how one can obtain such a representation from the expert features $F$. A first approach might be to put the expert features into bins to arrive again at discrete class labels, and then proceed with a standard contrastive loss function such as the pair-loss. Such an approach cannot generally lead to a bilipschitz representation; in particular, it cannot provide guarantees on $l_-$ and $l_+$ due to arbitrariness in choosing the bins and the absence of a relative distance measure between different bins. A similar problem occurs with other methods that generate pseudo-labels such as SleepPriorCL \cite{zhang2021sleeppriorcl}.

 As an alternative approach, one might add a linear layer $M=M_\phi$ on top of the encoder $E_\theta$ and jointly train $\theta$ and $\phi$ to predict the given expert features, e.g.\ by minimizing the MSE-loss. However, such a procedure does not necessarily lead to good guarantees for the two properties (P1), (P2) we are trying to fulfill, as the following shows:
\begin{proposition}\label{prop:prop1}
Let ${\mathcal L}_{mse}(f',f) = \|f'-f\|_2^2$ be the MSE-loss, $E_\theta$ the encoder, and $M_{\phi}:{\mathbb R}^{e} \to {\mathbb R}^{d}$ be a linear model. Then, even if $\theta$ and $\phi$ are such that ${\mathcal L}_{mse}(M_{\phi}\circ E_\theta(X),F)=0$, this does not provide any guarantees on $l_-,l_+$. When furthermore ${\rm dim}({\rm ker}(M_\phi)) > 0$ (in particular for $d>e$), a vanishing ${\mathcal L}_{mse}$ does not even guarantee $E_\theta$ to be a bilipschitz representation at all.
\end{proposition}

Prop.\ \ref{prop:prop1} shows that learning to predict the expert features does not provide any guarantees for the two properties (P1) or (P2). The proof can be found in App.\ \ref{sec:proof_prop_1}.

\subsection{Contrastive Learning with Continuous Features}\label{cl_conti_features}
The preceding section shows that two straightforward approaches do not guarantee an encoder $E$ to have the desired properties. We thus aim to find a new objective for CL which ensures the properties (P1) and (P2) to be fulfilled, and to a good degree at that. For this, we first propose to generalize the discrete similarity measure used within Eq.\ \ref{eq:pair_loss} to continuous labels or features. We do this by defining 
\begin{equation}\label{eq:vanilla_sim}
    s_{ij} := 1-\frac{\|f_i-f_j\|_2}{\max_{k,l}\|f_k-f_l\|_2}~,
\end{equation}
where the maximum can either be taken over the complete dataset $X$ or only over a subsample (batch). Other ways of defining the similarity measure are equally possible, see also the possibilities discussed in Sec.\ \ref{sec:practical_aspects}. We point out that, for discrete class labels which we assume here to be one-hot-encoded, our generalized similarity measure reduces to the discrete similarity measure $s_{ij}=\delta_{f_i=f_j}$ from Sec.\ \ref{sec:setting}. While this is not a necessary condition, it allows us to treat discrete and continuous features $f$ in a uniform manner with our generalized similarity measure. When plugging the new similarity (Eq.\ \ref{eq:vanilla_sim}) into the original pair-loss (Eq.\ \ref{eq:pair_loss}), the resulting loss function encourages the desired properties (P1) and (P2) to be fulfilled, as shown below. However, the resulting loss function has the issue of discontinuities in its derivative, similar to versions of the pair-loss (Eq.\ \ref{eq:pair_loss}) for continuous features (see also App.\ \ref{sec:loss_functions_cont_exp}), which might cause instabilities during optimization. We remedy this by designing a novel version of the continuous pair-loss, which we name the \emph{quadratic contrastive loss}:
\begin{equation}\label{eq:quadratic_con_loss}
    {\mathcal L}_{quad}(E(X),F) := \frac{1}{N^2}\sum_{i,j=1}^N\big((1-s_{ij})\Delta -D_{ij} \big)^2,
\end{equation}
where again $D_{ij} := \|E(x_i)-E(x_j)\|_2$. ${\mathcal L}_{quad}$ has the same minimum as the pair-loss (Eq.\ \ref{eq:pair_loss}), but possesses continuous derivatives w.r.t.\ $D_{ij}$ and thus $E$. Furthermore, in contrast to the procedure from Prop.\ \ref{prop:prop1}, minimizing ${\mathcal L}_{quad}$ does lead to (optimal) guarantees for $l_-$ and $l_+$:
\begin{proposition}\label{proposition_2}
Let ${\mathcal L}_{quad}$ be the quadratic contrastive loss (\ref{eq:quadratic_con_loss}) and $E_\theta$ the encoder. If $\theta$ is such that ${\mathcal L}_{quad}(E_\theta(X),F)=0$, then $E_\theta$ is a $[l_-,l_+]$-bilipschitz representation with $l_-=l_+=  \max_{i,j}\|f_i-f_j\|_2/\Delta $
\end{proposition}
While Prop.\ \ref{proposition_2} gives guarantees for the desired properties only on the training set $X$ (App.\ \ref{sec:proof_prop_2}), this can be boosted to statistical bounds for (P1) and (P2) on unseen test data (App.\ \ref{sec:statistical_bounds}). Such guarantees are relevant for downstream tasks such as outlier detection, searching, or (safe) active learning. 

\subsection{Implicit Hard-Negative Mining}\label{sec:hard_negative_mining}
In the previous section we introduced the quadratic contrastive loss ${\mathcal L}_{quad}$ (Eq.\ \ref{eq:quadratic_con_loss}), which puts equal weight on all pairs of datapoints. While this works well, the performance in practice often improves when higher weight is put on high-loss datapoint pairs. Such a strategy is known as ``hard-negative mining'' in CL \cite{wang2021understanding} and control-theory \cite{busseti2016risk}, and also improves our method (Sec.\ \ref{sec:ablation}). Rather than explicitly selecting the high-loss pairs, we perform \emph{implicit} hard-negative mining through a version of the softmax-function. This changes the loss to:
\begin{equation}\label{eq:expclr_loss}
     {\mathcal L}_{ExpCLR}^\tau(E(X),F) = \tau \log\left[\sum_{i,j=1}^N\frac{\exp\left(\frac{L_{ij}}{\tau}\right)}{N^2}\right],
\end{equation}
where $L_{ij} := \left((1-s_{ij})\Delta -D_{ij} \right)^2$ and $\tau\in {\mathbb R}^+$ is a temperature hyperparameter. The following proposition demonstrates that by changing $\tau$, one can control the strength of the implicit hard-negative mining:
\begin{proposition}\label{proposition_3} 

\begin{itemize}
    \item[]
    \item[(a)]In the limit $\tau \to 0$, minimizing ${\mathcal L}_{ExpCLR}^\tau(E(X),F)$ is equivalent to minimizing ${\mathcal L}_{max}(E(X),F) := \max_{i,j} L_{ij}$.
    \item[(b)]In the limit $\tau \to \infty$, minimizing ${\mathcal L}_{ExpCLR}^\tau(E(X),F)$ is equivalent to minimizing $ {\mathcal L}_{quad}(E(X),F)$.
\end{itemize}
\end{proposition}

Thus, the amount of hard-negative mining one wants to apply can be controlled by $\tau$ (proof in App.\ \ref{sec:proof_prop_3}). Similar limits exist for the NCE-Loss \cite{wang2021understanding}. A gradient-level analysis of the hard-negative mining loss can be found in App.\ \ref{sec:gradient_level_proof_hnm}. We refer to the method of minimizing the loss ${\mathcal L}_{ExpCLR}$ (Eq.\ \ref{eq:expclr_loss}), i.e.\ our quadratic contrastive loss with implicit hard-negative mining, as \emph{ExpCLR}.


\subsection{Practical Considerations for Contrastive Learning with Continuous Expert Features}\label{sec:practical_aspects}
Here we discuss two practical variations of ${\mathcal L}_{ExpCLR}$ we use in our experiments. First, while we introduced a simple similarity measure in Eq.\ \ref{eq:vanilla_sim}, in practice we use its square,
\begin{equation}\label{eq:our_sim}
     s_{ij} = \left(1-\frac{\|f_i-f_j\|_2}{\max_{k,l}\|f_k-f_l\|_2}\right)^2,
\end{equation}
due to its consistent superior performance in experiments. Other similarity measures, such as $ s_{ij} =\exp\left(-\|f_i-f_j\|^2_2/\sigma \right)$ \cite{kim2019deep} with a hyperparameter $\sigma$, would be equally possible. For an empirical comparison of the three similarity measures, see Sec.\ \ref{sec:ablation}.

Further, instead of using the unnormalized Euclidean metric and following \cite{kim2021embedding}, we normalize the Euclidean distance with $\mu_i=(1/N)\sum_j \|E(x_i)-E(x_j)\|_2$, i.e.\ we use $D_{ij} := \|E(x_i)-E(x_j)\|_2/\mu_i$. Alternative normalizations would be possible as well.

\section{Results}\label{sec:experiments}

\subsection{Datasets and Expert Features}
In the following we compare ExpCLR to several state-of-the-art methods on three real-world time-series datasets. 
We start by introducing the three datasets; see Tab.\  \ref{tab:dataset_description} for detailed specifications.

\begin{table}
\centering
\resizebox{0.485\textwidth}{!}{
\begin{tabular}{lllllll}
\toprule
{Dataset} & {\#Train} & {\#Test} & {Length} & {\#Exp.\ feat.} & {\#Chan.} & {\#Class.}\\
\midrule
HAR & 7352 & 2947 & 128 & 561 & 9 & 6  \\
SleepEDF & 35503 & 6805 & 3000 & 29 & 1 & 5  \\
Waveform & 59922 & 16645 & 2500 & 176 & 2 & 4  \\
\bottomrule
\end{tabular}
}
\caption{\textbf{Dataset Information:} Number of train and test samples, sample length $T$, dimension of expert features $d$, number of signal channels $c$, and number of classes $C$.}
\label{tab:dataset_description}
\end{table}

\textbf{Human Activity Recognition (HAR):}
The HAR dataset \cite{cruciani2019public} contains multi-channel sensor signals of 30 subjects, each performing one out of six possible activities. A Samsung Galaxy S2 device embeds accelerometers and gyroscopes which collected the data at a constant rate of 50Hz. In addition, the dataset already contains a 561-dimensional expert feature vector for each sample.

\textbf{Sleep Stage Classification (SleepEDF):}
In this classification task the goal is to classify five sleep stages from single-channel EEG signals, each sampled at 100Hz. The dataset originates from \cite{goldberger2000physiobank, 867928} and subjects are selected and preprocessed following previous studies \cite{eldele2021time}. We equip each signal with expert features computed from the time and frequency domain, as suggested and identified in \cite{HUANG2020105253}. 

\textbf{MIT-BIH Atrial Fibrillation (Waveform):}
This dataset \cite{goldberger2000physiobank} contains 23 long-term ECG recordings of humans suffering from atrial fibrillation. Two ECG signals are sampled at a constant rate of 250Hz and distinguish four different classes. We utilize the expert features designed by \cite{goodfellow2017classification} specifically for the artial fibrillation classification task.


To evaluate the expert features we report their resulting linear and KNN ($k=1$) classification accuracies in Tab.\ \ref{tab:unsupervised_results}.

\begin{table*}[t]
\centering
\resizebox{0.99\textwidth}{!}{
\begin{tabular}{lllllll}
\toprule
{Dataset} & \multicolumn{2}{l}{HAR} & \multicolumn{2}{l}{SleepEDF} & \multicolumn{2}{l}{Waveform} \\
{Performance (in \%)} &          Lin. Acc. &         KNN Acc. &          Lin.\ Acc. &         KNN Acc. &          Lin.\ Acc &         KNN Acc. \\
\midrule
Cross-Entropy (S) &  96.47 +/- 0.09 &  96.57 +/- 0.09 &  80.90 +/- 0.13 &  80.80 +/- 0.17 &  97.03 +/- 0.09 &  96.97 +/- 0.11 \\
Expert Features  &  96.01 +/- 0.00 &  87.90 +/- 0.00 &  77.00 +/- 0.00 &  73.50 +/- 0.00 &  44.40 +/- 0.00 &  92.00 +/- 0.00 \\
Random Init &  67.74 +/- 0.59 &  75.02 +/- 0.48 &  55.14 +/- 0.46 &  43.78 +/- 0.28 &  54.56 +/- 1.04 &  55.20 +/- 0.62 \\

\midrule
ExpCLR (U)  &  \textbf{91.18 +/- 0.41} &  88.72 +/- 0.22 &  \textbf{81.84 +/- 0.12} &  \textbf{74.82 +/- 0.12} &  \textbf{92.64 +/- 0.88} &  88.30 +/- 0.98 \\
SimCLR (U)&  90.70 +/- 0.30 &  \textbf{88.94 +/- 0.46} &  68.32 +/- 0.16 &  46.28 +/- 0.38 &  62.28 +/- 4.76 &  76.58 +/- 0.96 \\
SleepPriorCL (U) &  88.98 +/- 0.25 &  83.50 +/- 0.23 &  78.56 +/- 0.05 &  71.68 +/- 0.07 &  92.06 +/- 0.36 &  \textbf{88.70 +/- 0.64} \\
\citet{kim2021embedding} (U) &  89.02 +/- 0.15 &  86.72 +/- 0.31 &  75.70 +/- 0.25 &  61.12 +/- 0.27 &  83.96 +/- 1.75 &  81.80 +/- 0.93 \\

TS-TCC (U)&  90.57 +/- 0.15 &               -- &  80.68 +/- 0.24 &               -- &  82.17 +/- 2.53 &               -- \\
TREBA Contrastive Loss (U) &  78.40 +/- 1.79 &  65.90 +/- 0.16 &  77.73 +/- 0.63 &  70.20 +/- 0.08 &  90.53 +/- 0.68 &  81.13 +/- 0.52 \\

Expert Feature Decoding (U)&  85.20 +/- 2.69 &  79.83 +/- 3.01 &  80.73 +/- 0.05 & 74.97 +/- 0.45 &  91.83 +/- 1.10 &  82.70 +/- 3.80 \\

\bottomrule
\end{tabular}
}
\caption{\textbf{Unsupervised Learning Comparison:} Comparison of ExpCLR to state-of-the-art unsupervised representation learning methods on the HAR, SleepEDF and Waveform datasets. The table shows the mean performance and standard error over five independent trials for the linear and KNN ($k=1$) classification accuracies. For better comparison we also include the performance of the representations obtained from supervised learning (full labeled data), of the expert features, and of the randomly initialized encoder network. Overall, ExpCLR outperforms the other unsupervised methods and even surpasses the supervised performance on the SleepEDF dataset.}
\label{tab:unsupervised_results}
\end{table*}

\subsection{Implementation Details -- Model and Training}\label{sec:imp_details}
Next, we briefly present our model architecture and how we train our models in each setting; more details can be found in App.\ \ref{sec:imp_details_app}.
To capture relevant temporal properties and to improve training stability \cite{bai2018empirical}, we choose as a base encoder temporal convolutional network (TCN) \cite{lea2017temporal} layers in a ResNet \cite{he2016deep} architecture with eight such temporal blocks. Note that ExpCLR is not restricted to this architecture. To reach a pre-defined embedding dimensionality we add a two-layer fully connected neural network on top of the ResNet base encoder to arrive at our backbone encoder network. 

In our work we consider three different modes of training: 
\begin{enumerate}
    \item  The \emph{unsupervised} (U) training mode, where the encoder is optimized with the respective contrastive loss function on the input data time-series $X$ and the expert features $F$ only.
    
    \item The \emph{supervised} (S) training mode, where the encoder is trained with either a supervised contrastive loss, which uses labels $y_i$ instead of expert features $f_i$, or with the cross-entropy loss function on the input time-series $X$ and the whole or part of the labels $Y$. 
    
    \item The \emph{semi-supervised} (SS) training mode, where the encoder is first trained with the unsupervised training mode on the whole training set $(X,F)$ and then this pretrained encoder is fine-tuned with a supervised training step on some percentage of the labels $Y$. 
\end{enumerate}

While for hyperparameter optimization we split the training set $X$ into $80\%$ training and $20\%$ validation data, for our comparisons experiments we make use of the full training set and evaluate the representations on the test set. 
The number of epochs for each dataset is selected such that all algorithms are able converge. For the optimization step we used the Adam optimizer with parameters $\beta_1 = 0.9$,  $\beta_2 = 0.999$ and exponential decay $\gamma = 0.99$.
To enable a fair comparison between ExpCLR and the competing methods, we optimize the learning rate for each method and dataset individually via a grid search and identify $\tau = 1$, $\Delta = 1$ (Eq.\ \ref{eq:expclr_loss}), embedding dimension $e = 100$ and batch size of 64 as a good compromise over all datasets and algorithms. For more information on model- and loss-specific parameters, see Sec. \ref{sec:ablation}, App. \ref{sec:sens_ana} and App.\ \ref{sec:imp_details_app}.


To verify the goodness of our representations, we evaluate two kinds of classifiers on the representation: 
Linear classifiers perform well for representations where all classes are linearly separable. Second, KNN classifiers can even perform well when classes are not linearly separable, but tend to perform worse for clusters that are not separated by a large margin; this problem is most apparent for small $k$. 
We thus use the performance difference of the linear and KNN ($k=1$) classifier to investigate how well different classes are mapped into individual well-separated clusters.  

\begin{figure*}[ht]
    \centering
    \includegraphics[scale=0.675]{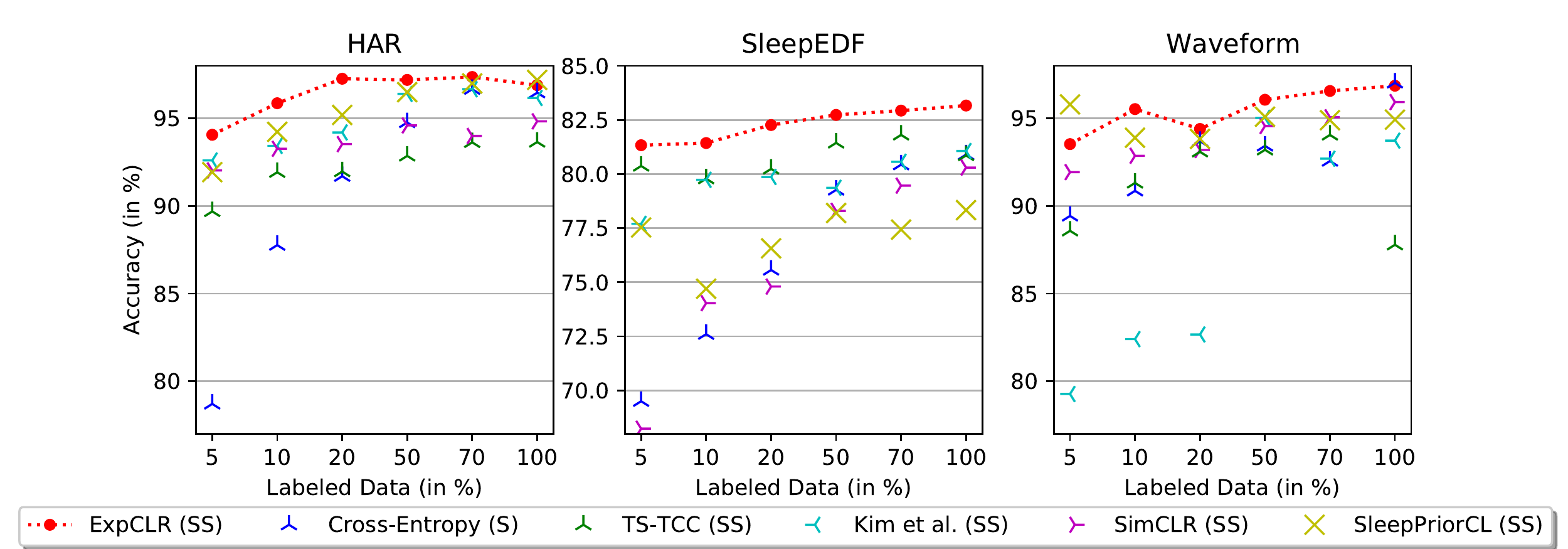}
    \caption{\textbf{Semi-Supervised Learning Comparison:} Comparison of ExpCLR (SS) to semi-supervised representation learning methods on the HAR, SleepEDF, and Waveform datasets. Shown are the mean linear classification accuracies for different percentages of labeled data available during the supervised fine-tuning step.  Across all datasets, ExpCLR outperforms all other methods. Further, while ExpCLR achieves consistent performance on all three datasets, all competing methods have drastically varying performance across datasets.
    }
    \label{fig:label_ratios_linear_new}
\end{figure*}

\subsection{Competing Methods}\label{sec:competing_methods}                           
For the unsupervised comparison of ExpCLR to other methods, we distinguish two different groups of CL algorithms. The first group consist of algorithms, which use transformations of the input data to create positive samples. Here we compare to SimCLR \citep{chen2020simple} and further to TS-TCC \cite{eldele2021time}, that combines classical CL with time-series forecasting to learn representations.
For SimCLR we tested a range of different augmentations and found scaling and dropout to work the best, while for TS-TCC we employ weak and strong augmentations. Further details can be found in App.\ \ref{sec:comp_details}.
The second group includes methods which also use expert features. SleepPriorCL \citep{zhang2021sleeppriorcl} and the contrastive loss used by TREBA \citep{sun2021task} both create expert feature based pseudo-labels via some form of discretization, which are then used with a version of the supervised contrastive loss introduced by \cite{khosla2020supervised}. Another method in this group is introduced by \cite{kim2021embedding}, a state-of-the-art metric learning method. It aims to achieve the same goal as we do and try to pull similar points closer together \cite{kim2021embedding}, while pushing dissimilar ones further apart. We simply replace their teacher model output with our expert features to be comparable. Lastly, we also compare to the embedding learned by Expert Feature Decoding: Here, the embedding is given by the output of the penultimate layer of a network that is trained by learning to predict the expert features from the input time-series \citep{sun2021task}. During training we minimized the MSE-loss and add a projection layer to the architecture used by the other methods. Expert Feature Decoding is used as part of the TREBA-objective \citep{sun2021task} and is discussed theoretically in Sec.\ \ref{sec:fail_feed_forward}.

For the supervised fine-tuning step we use the approach described in the original works or utilize the natural extension for each algorithm. For SimCLR we use supervised CL \cite{khosla2020supervised} and for \cite{kim2021embedding} we use the pair-Loss \citep{hadsell2006dimensionality,le2020contrastive}, because this is the loss that it naturally reduces to for class labels. Further, as ExpCLR allows to simply replace expert features with labels in order to perform a supervised fine-tuning step or a fully supervised training, we do this.

We selected the competing methods to cover a broad range of algorithms. All methods can be considered state-of-the-art in their respective domains. We implemented all methods except for TS-TCC inside our repository. 

\subsection{Comparison for Unsupervised Representation Learning}
In this section we compare the performance of ExpCLR against several state-of-the-art unsupervised representation learning methods, using a linear and a KNN ($k=1$) classifier on top of the learned embedding. We compare ExpCLR on all three datasets against SimCLR, TS-TCC, SleepPriorCL, TREBA Contrastive Loss, Expert Feature Decoding and \cite{kim2021embedding}. In addition, we use the randomly initialized encoder network performance (Random Init), an encoder trained with a supervised cross-entropy loss, using the full labeled dataset, and the performance on the expert features themselves as baseline comparisons. The results of the comparison are shown in Tab.\ \ref{tab:unsupervised_results}.

The superior performance of ExpCLR can be clearly seen, as we only perform on par with SimCLR on HAR and with SleepPriorCL on Waveform w.r.t.\ KNN ($k=1$) accuracy. ExpCLR also shows much higher consistency across all datasets, while most other algorithms significantly underperform on at least one of the datasets. Further, we are even able to exceed the expert feature performance on at least one performance metric on all three datasets. This could indicate that ExpCLR is able to learn new additional features from the raw time-series data on top of the provided expert features. In addition, we can even surpass the supervised performance on the SleepEDF dataset. This underlines how powerful the approach of ExpCLR.

\begin{figure*}[t]
    \centering
    \includegraphics[scale=0.6]{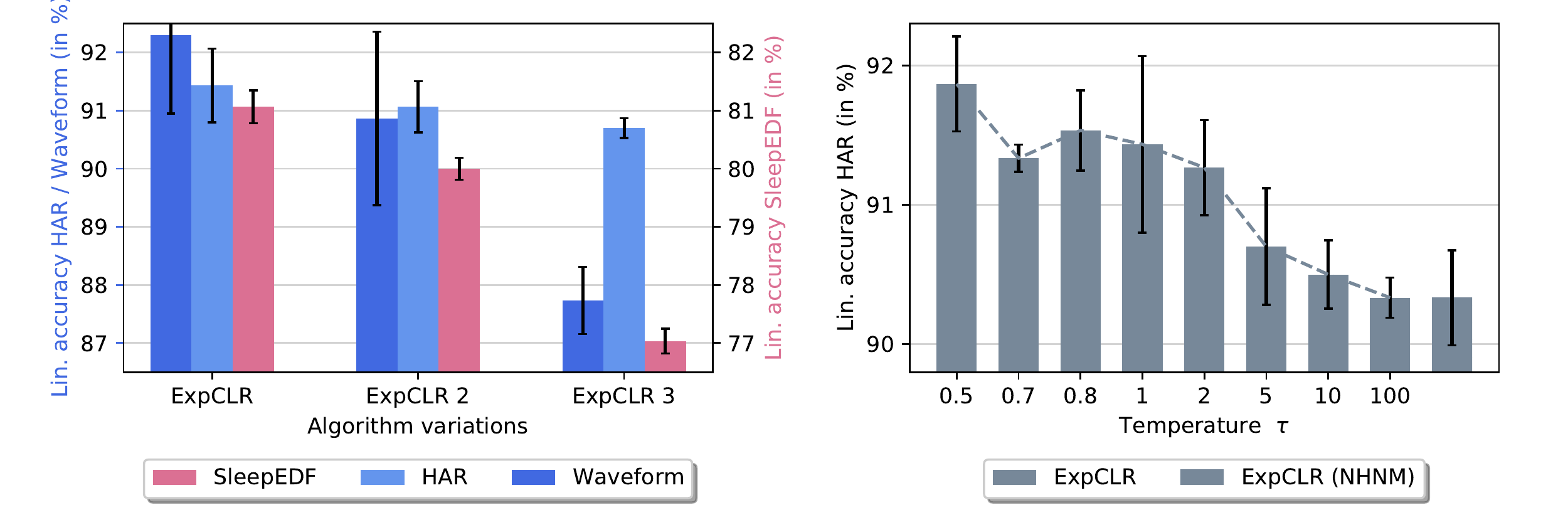}
    \caption{\textbf{Ablation Experiments:} The left panel shows a comparison of three different similarity measures in ExpCLR in the unsupervised training mode. While ExpCLR uses the similarity measure (Eq.\ \ref{eq:our_sim}), ExpCLR 2 uses the similarity (Eq.\ \ref{eq:vanilla_sim}), and  ExpCLR 3 uses the exponential similarity introduced by \cite{kim2021embedding} (Sec.\ \ref{sec:practical_aspects}). Our choice clearly outperforms the other ones. 
    The right panel investigates the effectiveness of our implicit hard-negative mining scheme (Sec.\ \ref{sec:hard_negative_mining}) for the unsupervised setting on the HAR dataset. We plot the linear accuracies over different values of $\tau$ and also compare these to a version of ExpCLR without hard-negative mining (NHNM), i.e.\ the quadratic contrastive loss (Eq.\ \ref{eq:quadratic_con_loss}). The results show that hard-negative mining can improve the performance significantly. 
    }

    \label{fig:ablation_loss_tau}
\end{figure*}
\subsection{Comparison for Semi-Supervised Representation Learning}\label{exp:semi-supervised}
Here, we compare ExpCLR to the competing methods in the semi-supervised setting. In this setting we fine-tune a representation, learned in the unsupervised setting, using some fraction of the full labeled data. 
The results of the comparison for the HAR, SleepEDF and Waveform datasets for labeled data ratios of $5\%, 10\%, 20 \%, 50\%, 70\%$ and $100\%$ are shown in Fig.\ \ref{fig:label_ratios_linear_new} (see App.\ \ref{sec:label_ratio_knn} for the KNN accuracy).

ExpCLR consistently surpasses the accuracy of all competing methods across all datasets over different label percentages. Further, using only  $20\%$ labeled data ExpCLR (SS) is able to outperform any competing method with any amount of labeled data, even for $ 100\% $, on the HAR and SleepEDF datasets. In addition, ExpCLR's unsupervised pretraining
increases the label efficiency significantly, since ExpCLR is able to sustain its performance attained on the fully labeled dataset up to a minor decrease in accuracy of less than $3.5 \%$ for the lowest labeled data percentage. In contrast, the representations learned by supervised cross-entropy drop by more than $17.5 \%$. 

The consistent superior performance of ExpCLR across all datasets is noteworthy as it does not use any data transformations. Further, the comparison clearly shows the superior performance of our loss function compared to \citet{kim2021embedding}. Another interesting observation is that almost all methods that make use of an unsupervised pretraining outperform supervised methods for low percentages of labeled data. This underlines the importance and effectiveness of finding good approaches that make use of unlabeled data, as ours. All numerical values of Fig.\ \ref{fig:label_ratios_linear_new} incl.\ variances can be found in App.\ \ref{sec:tab_data_lin} (see also App.\ \ref{sec:tab_data_knn}).

\subsection{Ablation Studies}\label{sec:ablation}

To guide the design of our ExpCLR method we conducted several ablation studies. The first one investigates the choice of similarity measure in ExpCLR and is shown in the left panel of Fig.\ \ref{fig:ablation_loss_tau}. 
We compare both similarities (Eq. \ref{eq:vanilla_sim}) and (Eq. \ref{eq:our_sim}) from our methods section and the Gaussian similarity (Sec.\ \ref{sec:practical_aspects}) introduced by \citet{kim2021embedding} in the unsupervised setting for all three real-world datasets. The results show that the quadratic similarity measure (Eq. \ref{eq:our_sim}) outperforms the other similarity measures, therefore justifying our choice.

The second ablation study investigates the effectiveness of the hard-negative mining strategies introduced in Sec.\ \ref{sec:hard_negative_mining}, which ExpCLR employs. The right panel of Fig.\ \ref{fig:ablation_loss_tau} shows the unsupervised accuracy of ExpCLR for several values of the temperature $\tau$ and also compares to ExpCLR without hard-negative mining (NHNM). Using the insights gained from Prop.\ \ref{proposition_3} (a), Fig.\ \ref{fig:ablation_loss_tau} shows that stronger hard-negative mining (decreasing $\tau$) can improve the performance.
Further, as shown theoretically in Prop.\ \ref{proposition_3} (b), the results nicely demonstrate that the performance convergences to ExpCLR (NHNM) for large values of $\tau$. More ablation studies and a sensitivity analysis for the hyperparameter $\Delta$, the batch-size and the dimension of the embedding can be found in the appendix (see App.\ \ref{sec:sens_ana}).


\section{Discussion}\label{sec:discussion}
In this paper we introduce ExpCLR, a novel contrastive representation learning algorithm that can utilize continuous or discrete expert features. We first propose two properties a useful time-series representation should fulfill. In a second step, we design ExpCLR to be applicable in the unsupervised and semi-supervised domains and show that the loss function we devised for ExpCLR leads to a representation that encourages both properties. We demonstrate on an array of experiments the superior performance of ExpCLR compared to state-of-the-art methods, sometimes exceeding their accuracies despite using only a fraction of the labels.

We see in ExpCLR an alternative to the classical transformation-based contrastive learning (CL) approaches, as ExpCLR does not make use of any transformations.
Nevertheless, ExpCLR is able to outperform these classical approaches in both the un- and semi-supervised settings. This is especially noteworthy for datasets where domain experts can provide valuable expert features. Thus, ExpCLR is applicable to any dataset for which expert features are available, and is not limited to time-series datasets. 
In addition, ExpCLR can also be applied to supervised CL with datasets containing continuous labels, e.g.\ regression tasks such as pose estimation. 

We envision our ExpCLR approach to not only serve as a standalone method for representation learning, but to be applicable to any task or dataset where (continuous) expert features are available in order to infuse expert knowledge into neural networks. Apart from pretraining, this might be achieved by jointly training our ExpCLR loss together with a task-specific loss function , as done by TREBA \citep{sun2021task}, which should increase the task performance while being more label-efficient. 

\bibliography{icml2022}
\bibliographystyle{icml2022}

\newpage

\appendix

\onecolumn

\begin{center}
    {\Large{Supplementary Material for}}
    
    {\LARGE{\textbf{Utilizing Expert Features for Contrastive Learning}}}
    
    {\LARGE{\textbf{of Time-Series Representations}}}
    
\end{center}

\smallskip

\section{Additional Experiments}\label{sec:add_exp}

\subsection{Figure with KNN $(k=1)$ Classification Results for the Semi-Supervised Experiments}\label{sec:label_ratio_knn}
\begin{figure*}[h]
    \centering
    \includegraphics[scale=0.675]{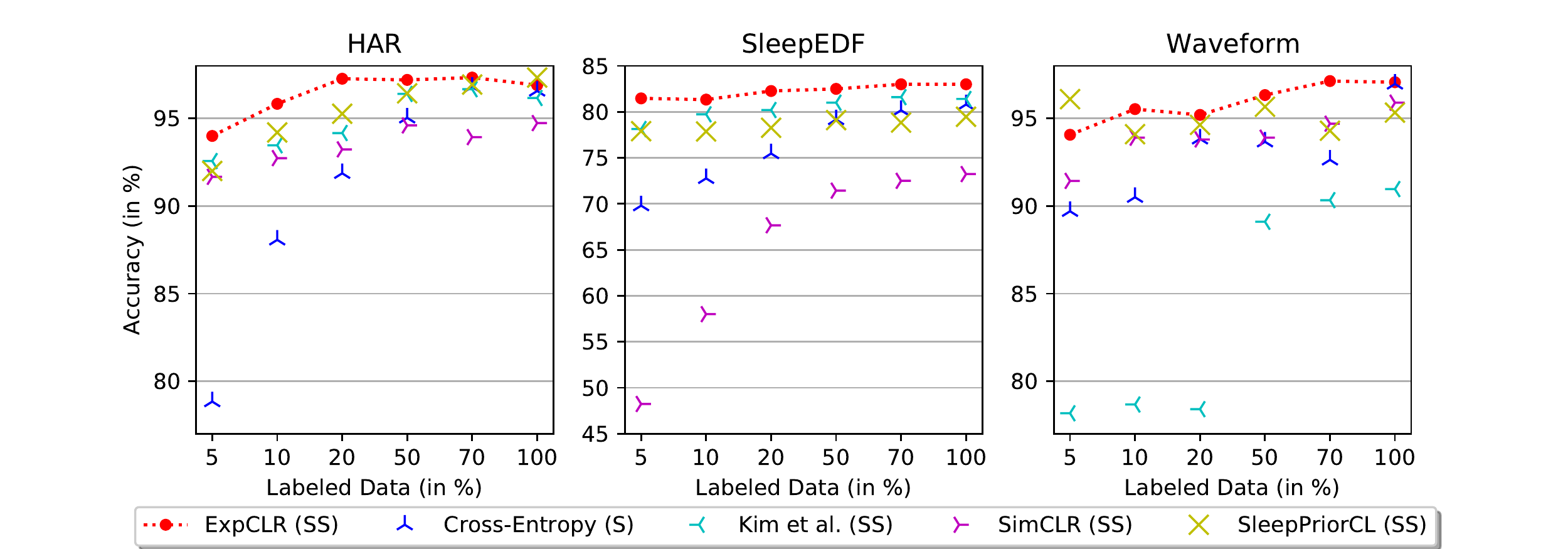}
    \caption{\textbf{Semi-Supervised Comparison KNN ($k=1$):} The figure shows the same setting as Fig.\ \ref{fig:label_ratios_linear_new}, but instead of the linear classification accuracy all methods are evaluated with the KNN ($k=1$) classification accuracy.}
    \label{fig:label_ratios_linear_knn}
\end{figure*}

See Fig.\ \ref{fig:label_ratios_linear_knn} for a comparison of the KNN ($k=1$) accuracies on our three real-world datasets (analogous to Fig.\ \ref{fig:label_ratios_linear_new} for the linear classification accuracy). For all numerical values including standard deviations, see App.\ \ref{sec:tab_data_knn}.

\subsection{Sensitivity Analysis}\label{sec:sens_ana}
We perform sensitivity analysis for $\Delta$ from Eq.\ \ref{eq:quadratic_con_loss}, batch size during training, and dimension of $e$ the embedding space to which the input samples are mapped. To analyse the sensitivity, we train ExpCLR in unsupervised training mode and evaluate the linear accuracy of the embedding space. The results are shown in Fig.\ \ref{fig:abl_delta_batch_emb}. One can see that:

\begin{figure*}[h]
    \centering
    \includegraphics[scale=0.7]{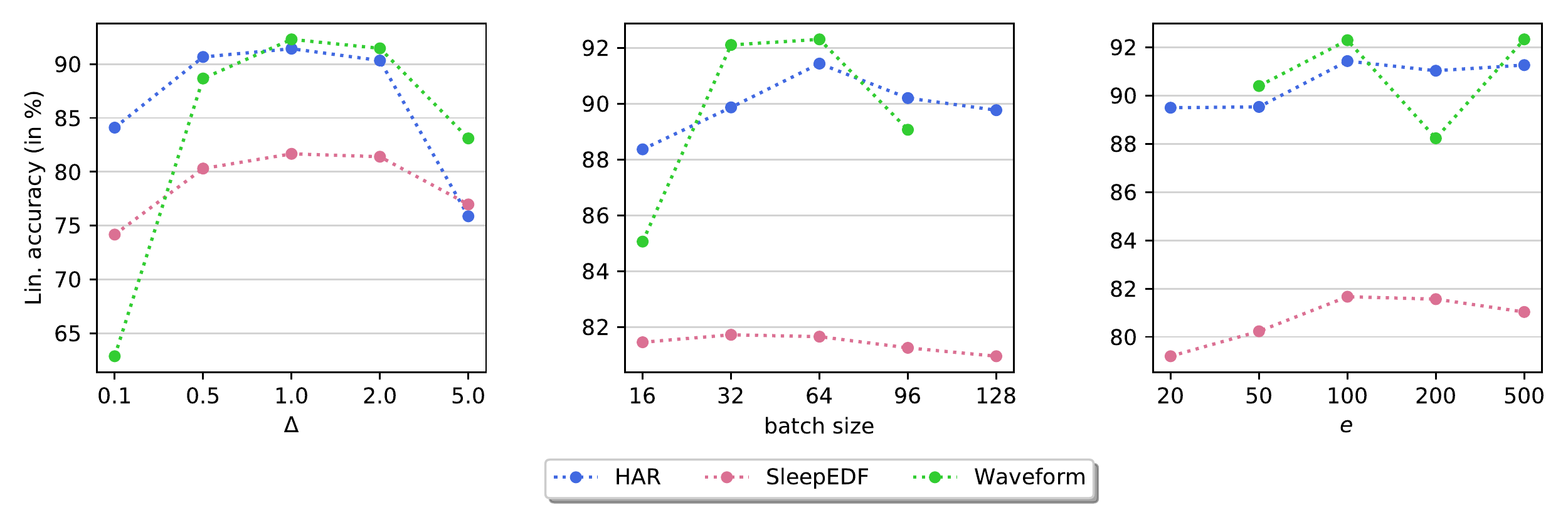}
    \caption{\textbf{Sensitivity Analysis ExpCLR Unsupervised:} The figure shows three sensitivity analyses for ExpCLR unsupervised training, evaluated by linear accuracy. The left panel shows results for different $\Delta$ in Eq.\ \ref{eq:expclr_loss}, in the middle panel for different batch sizes, and in the right panel for different embedding dimensions $e$.}
    \label{fig:abl_delta_batch_emb}
\end{figure*}

\begin{itemize}
    \item \textbf{Left plot}: There is an optimal value for $\Delta$ which is robust against modifications within a certain range. For too low or too high $\Delta$, linear accruacy tends to deteriorate because pushing the embedding to those margins can result in more difficult optimization leading to training instability. For all contrastive learning approaches that include this hyperparameter $\Delta$ in their losses, we choose $\Delta = 1$ for all datasets.
    
    \item \textbf{Middle plot}: For small batch sizes, performance drops as there might not be enough dissimilar samples within the same batch \cite{chen2020simple}. Also for too large batch sizes the performance decreases for all datasets consistently. For all algorithms and experiments we choose the $batch size = 64$.
    
    \item \textbf{Right plot}: Higher embedding dimensions tend to improving accuracys because the placement of embedding vectors and their separation might be easier in higher dimensional space. Note that although a low-dimensional embedding is generally preferable, too low dimensions may cause unstable training as different input vectors are forced to be pushed towards similar representations which goes against the nature of the loss function from Eq. \ref{eq:expclr_loss}. We observe an embedding size of $e=100$ as applicable and choose it over all datasets and algorithms consistently.
\end{itemize}

\subsection{Comparison ExpCLR (S) vs.  ExpCLR (SS)}\label{sec:comp_sup_vs_semi}

\begin{figure*}[h]
    \centering
    \includegraphics[scale=0.6]{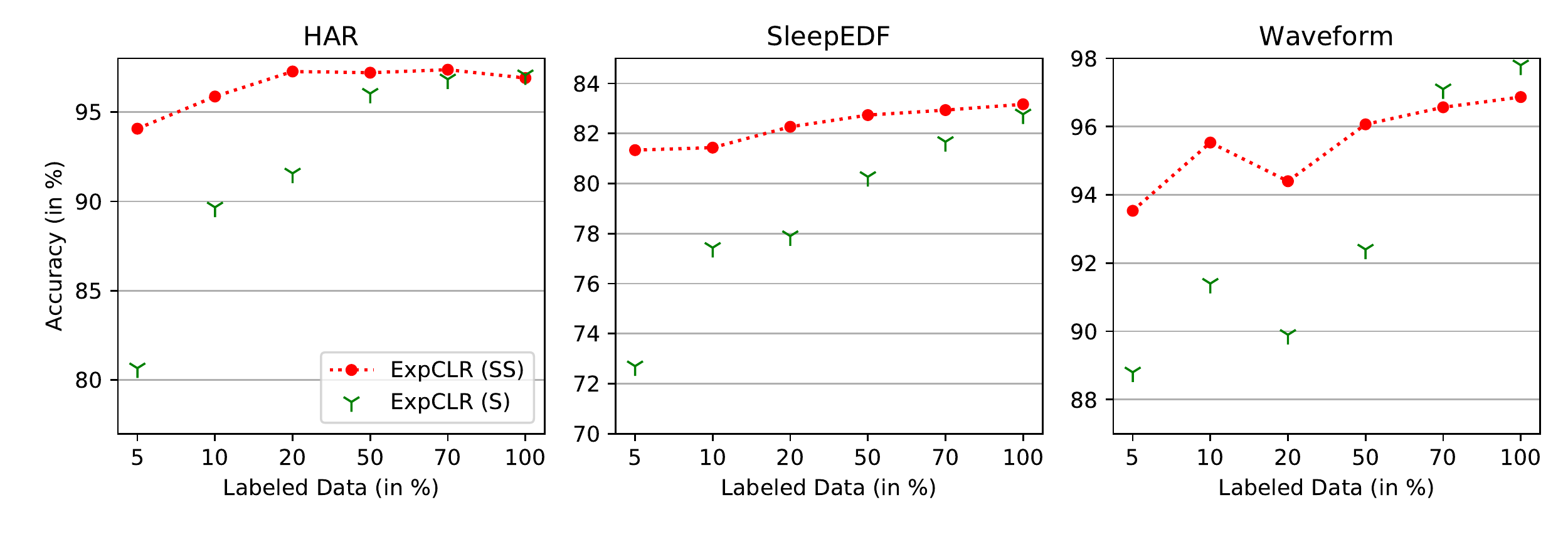}
    \caption{\textbf{ExpCLR (SS) vs ExpCLR (S) on Linear Classification Accuracies:} Shown is the same setting as in Fig.\ \ref{fig:label_ratios_linear_new} but only comparing the ExpCLR method in the semi-supervised and supervised settings. The results show that the unsupervised pretraining enables ExpCLR (SS) to clearly surpass the performance obtained by ExpCLR (S) overall, especially for lower percentages of labeled data.}
    \label{fig:comp_quadexp_expclr}
\end{figure*}

\begin{figure*}[h]
    \centering
    \includegraphics[scale=0.6]{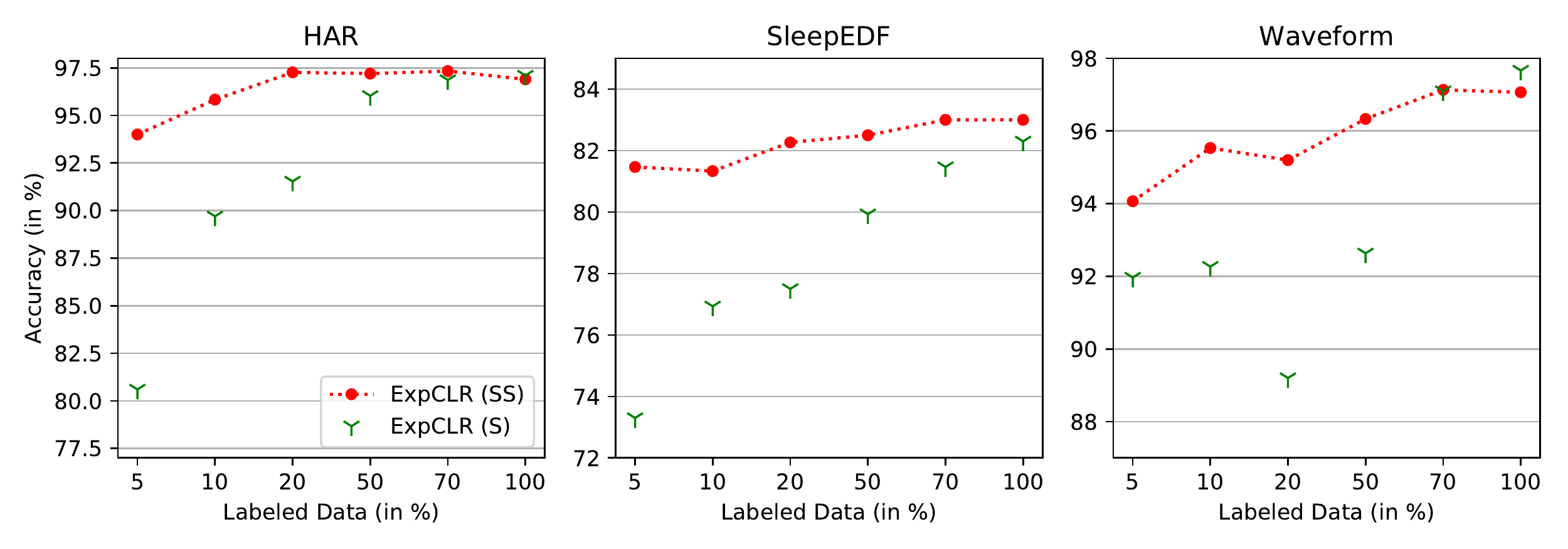}
    \caption{\textbf{ExpCLR (SS) vs ExpCLR (S) on KNN ($k=1$) Classification Accuracies:} The figure shows the same comparison as shown in Fig.\ \ref{fig:comp_quadexp_expclr} but instead of the linear accuracies it shows uses  KNN ($k=1$) accuracies to evaluate the quality of the representations. The results complement the findings of Fig.\ \ref{fig:comp_quadexp_expclr}.}
    \label{fig:comp_quadexp_expclr_knn}
\end{figure*}

Figs.\ \ref{fig:comp_quadexp_expclr} and \ref{fig:comp_quadexp_expclr_knn} show a comparison of the supervised (S) and semi-supervised (SS) versions of ExpCLR for the linear and KNN ($k=1$) classification accuracies they achieve, plotted over the fraction of labeled data used.

\subsection{Comparison of ExpCLR (SS) vs. Competing Methods Using ExpCLR (S) for the Fine-Tuning Step}\label{sec:comp_all_alg_with_sup_expclr}
\begin{figure*}[h]
    \centering
    \includegraphics[scale=0.6]{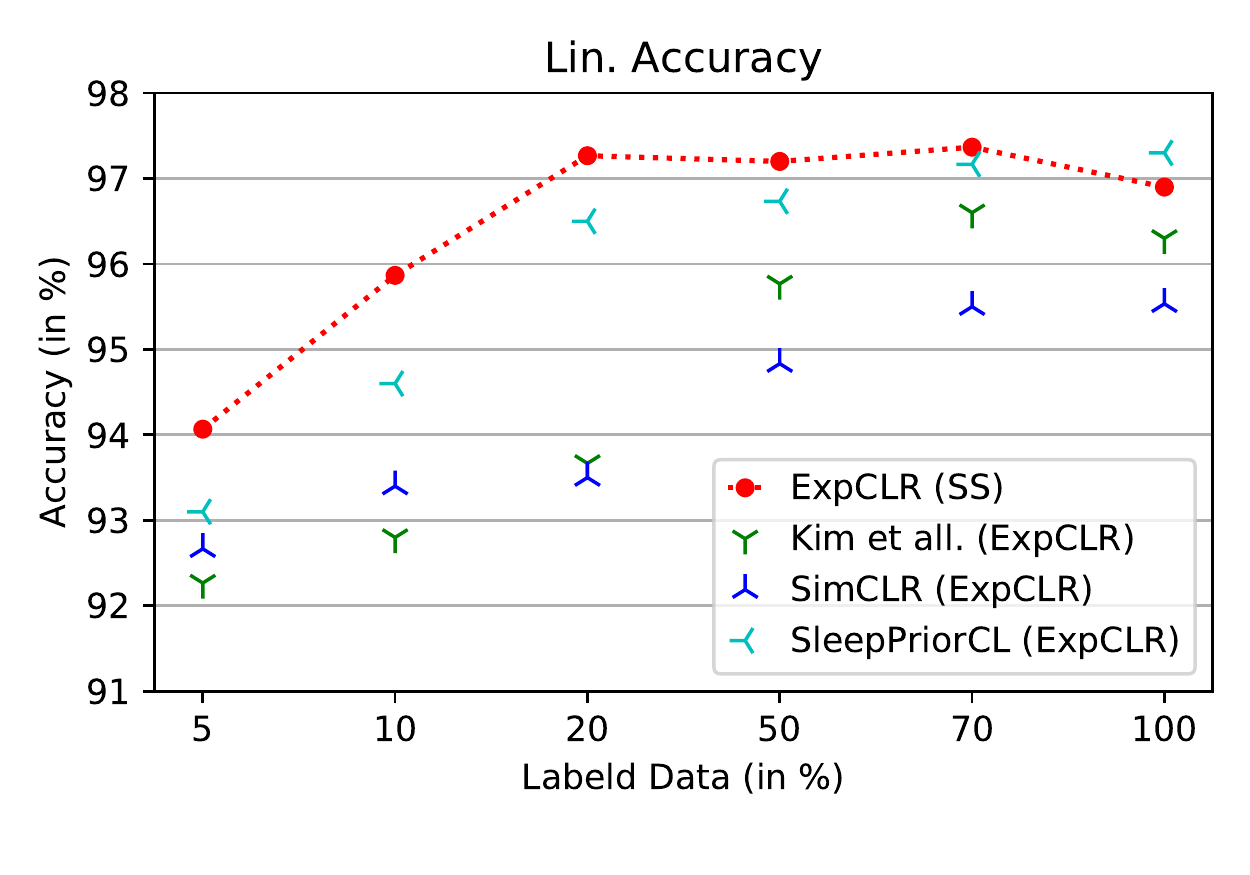}
    \caption{\textbf{Comparisons of ExpCLR (SS) vs Competing Methods with ExpCLR (S) Fine-Tuning}: The figure shows a comparison on the HAR dataset of ExpCLR (SS) with the competing methods but instead of using the standard fine-tuning step for each step, we apply ExpCLR (S) as a fine-tuning step for each competing method. The results indicate that while ExpCLR (S) performs well as a fine-tuning method, a key ingredient for the good performance of ExpCLR (SS) is the pretraining step with ExpCLR (U), using expert features.}
    \label{fig:comp_ft_all_expclr}
\end{figure*}

See Fig.\ \ref{fig:comp_ft_all_expclr} for a comparison of ExpCLR (SS) with the competing methods, where instead of using the standard fine-tuning step for each method (as was shown in Fig.\ \ref{fig:label_ratios_linear_new} and described in Sec.\ \ref{sec:competing_methods}) we apply ExpCLR (S) as the fine-tuning step for each competing method. 

\subsection{Tabular Data of the Linear Classification Accurracies for the Semi-Supervised Experiments}\label{sec:tab_data_lin}
The tables Tab.\  \ref{tab:mean_var_semi_har}, \ref{tab:mean_var_semi_sleep} and \ref{tab:mean_var_semi_waveform} show the tabluar data for Fig.\ \ref{fig:label_ratios_linear_new} (Sec.\ \ref{exp:semi-supervised}).

\begin{table}[h]
\centering
\resizebox{0.99\textwidth}{!}{
\begin{tabular}{lllllll}
\toprule
{Labeled Data} &              5\% &             10\% &             20\% &             50\% &             70\% &            100\% \\
\midrule
ExpCLR (SS) &  \textbf{94.07 +/- 0.71} &  \textbf{95.87 +/- 0.38} &  \textbf{97.27 +/- 0.54} &  \textbf{97.20 +/- 0.37} &  \textbf{97.37 +/- 0.27} &  96.90 +/- 0.29 \\
ExpCLR (S) &  80.67 +/- 2.65 &  89.67 +/- 0.31 &  91.57 +/- 0.83 &  96.03 +/- 0.26 &  96.83 +/- 0.15 &  97.07 +/- 0.49 \\
Cross-Entropy (S) &  78.70 +/- 0.90 &  87.77 +/- 0.19 &  91.70 +/- 0.50 &  94.77 +/- 0.58 &  96.67 +/- 0.10 &  96.47 +/- 0.12 \\
TS-TCC (SS) &  89.69 +/- 0.62 &  91.93 +/- 0.18 &  91.97 +/- 0.21 &  92.86 +/- 0.05 &  93.63 +/- 0.25 &  93.65 +/- 0.39 \\
\citet{kim2021embedding} (SS) &  92.60 +/- 0.64 &  93.43 +/- 0.93 &  94.20 +/- 0.87 &  96.40 +/- 0.08 &  96.67 +/- 0.26 &  96.17 +/- 0.18 \\
SimCLR (SS) &  92.03 +/- 0.59 &  93.27 +/- 0.41 &  93.53 +/- 0.27 &  94.60 +/- 0.08 &  94.00 +/- 0.29 &  94.83 +/- 0.15 \\
SleepPriorCL (SS) &  91.93 +/- 0.57 &  94.23 +/- 0.47 &  95.20 +/- 0.75 &  96.50 +/- 0.37 &  97.00 +/- 0.28 &  \textbf{97.20 +/- 0.26} \\
\bottomrule
\end{tabular}
\caption{Tabular data for the semi-supervised comparison shown in Fig.\ \ref{fig:label_ratios_linear_new} for linear classification accuracies on the HAR dataset. }
\label{tab:mean_var_semi_har}
}
\end{table}

\begin{table}[h]
\centering
\resizebox{0.99\textwidth}{!}{
\begin{tabular}{lllllll}
\toprule
{Labeled Data} &              5\% &             10\% &             20\% &             50\% &             70\% &            100\% \\
\midrule
ExpCLR (SS) &  \textbf{81.33 +/- 0.07} &  \textbf{81.43 +/- 0.10} &  \textbf{82.27 +/- 0.20} &  \textbf{82.73 +/- 0.03} &  \textbf{82.93 +/- 0.07} &  \textbf{83.17 +/- 0.18} \\
ExpCLR (S) &  72.70 +/- 0.34 &  77.43 +/- 0.21 &  77.90 +/- 0.52 &  80.27 +/- 0.24 &  81.67 +/- 0.35 &  82.77 +/- 0.05 \\
Cross-Entropy (S) &  69.50 +/- 0.33 &  72.60 +/- 0.71 &  75.57 +/- 0.07 &  79.27 +/- 0.30 &  80.43 +/- 0.05 &  80.90 +/- 0.17 \\
 TS-TCC (SS) &  80.37 +/- 0.22 &  79.79 +/- 0.08 &  80.24 +/- 0.15 &  81.44 +/- 0.18 &  81.81 +/- 0.15 &  80.86 +/- 0.37 \\
\citet{kim2021embedding} (SS) &  77.70 +/- 0.50 &  79.73 +/- 0.03 &  79.87 +/- 0.32 &  79.37 +/- 0.03 &  80.57 +/- 0.72 & 81.07 +/- 0.34 \\
SimCLR (SS) &  68.23 +/- 0.05 &  74.03 +/- 0.30 &  74.80 +/- 0.42 &  78.30 +/- 0.26 &  79.47 +/- 0.24 &  80.30 +/- 0.14 \\
SleepPriorCL (SS) &  77.53 +/- 0.12 &  74.70 +/- 0.70 &  76.57 +/- 0.67 &  78.20 +/- 0.22 &  77.43 +/- 0.43 &  78.33 +/- 0.30 \\
\bottomrule
\end{tabular}

\caption{Tabular data for the semi-supervised comparison shown in Fig.\ \ref{fig:label_ratios_linear_new} for linear classification accuracies on the SleepEDF dataset.}
\label{tab:mean_var_semi_sleep}
}
\end{table}

\begin{table}[h]
\centering
\resizebox{0.99\textwidth}{!}{
\begin{tabular}{lllllll}
\toprule
{Labeled Data} &              5\% &             10\% &             20\% &             50\% &             70\% &            100\% \\
\midrule
ExpCLR (SS) &  93.53 +/- 0.45 &  \textbf{95.53 +/- 0.88} & \textbf{ 94.40 +/- 1.94} &  \textbf{96.07} +/- 1.07 &  96.57 +/- 0.71 &  96.87 +/- 0.64 \\
ExpCLR (S) &  88.80 +/- 3.87 &  91.40 +/- 2.87 &  89.90 +/- 1.14 &  92.40 +/- 1.34 &  \textbf{97.10 +/- 0.45} &  \textbf{97.80 +/- 0.08} \\
Cross-Entropy (S) &  89.43 +/- 1.95 &  90.87 +/- 1.20 &  93.70 +/- 0.54 &  93.43 +/- 0.79 &  92.57 +/- 2.97 &  97.03 +/- 0.12 \\
TS-TCC (SS) &  88.58 +/- 2.18 &  91.32 +/- 0.58 &  93.09 +/- 0.40 &  93.21 +/- 0.87 &  94.06 +/- 0.30 &  87.79 +/- 1.52 \\
\citet{kim2021embedding} (SS) &  79.27 +/- 5.28 &  82.40 +/- 6.53 &  82.67 +/- 4.23 &  95.03 +/- 1.19 &  92.70 +/- 2.42 &  93.73 +/- 1.58 \\
SimCLR (SS) &  91.93 +/- 0.76 &  92.87 +/- 0.84 &  93.20 +/- 1.07 &  94.57 +/- 0.73 &  95.07 +/- 1.05 &  95.93 +/- 0.97 \\
SleepPriorCL (SS) &  \textbf{95.80 +/- 0.87} &  93.90 +/- 1.41 &  93.83 +/- 0.90 &  95.10 +/- 0.86 &  94.90 +/- 0.99 &  94.93 +/- 0.83 \\
\bottomrule
\end{tabular}

\caption{Tabular data for the semi-supervised comparison shown in Fig.\ \ref{fig:label_ratios_linear_new} for linear classification accuracies on the Waveform dataset.}
\label{tab:mean_var_semi_waveform}
}
\end{table}

\subsection{Tabular Data of the KNN ($k=1$) Classification Accuracies for the Semi-Supervised Experiments}\label{sec:tab_data_knn}

The tables Tab.\ \ref{tab:mean_var_semi_har_knn}, \ref{tab:mean_var_semi_sleep_knn} and \ref{tab:mean_var_semi_waveform_knn} show the tabluar data for Fig.\ \ref{fig:label_ratios_linear_knn} (App.\ \ref{sec:label_ratio_knn} and Sec.\ \ref{exp:semi-supervised}).

\begin{table}[h]
\centering
\resizebox{0.99\textwidth}{!}{
\begin{tabular}{lllllll}
\toprule
{Labeled Data} &              5\% &             10\% &             20\% &             50\% &             70\% &            100\% \\
\midrule
ExpCLR (SS) &  \textbf{94.00 +/- 0.70} &  \textbf{95.83 +/- 0.38} &  \textbf{97.27 +/- 0.54} &  \textbf{97.20 +/- 0.37} &  \textbf{97.33 +/- 0.24} &  96.90 +/- 0.29 \\
ExpCLR (S) &  80.60 +/- 2.58 &  89.70 +/- 0.31 &  91.53 +/- 0.87 &  96.03 +/- 0.26 &  96.87 +/- 0.14 &  97.10 +/- 0.46 \\
Cross-Entropy (S) &  78.83 +/- 0.94 &  88.07 +/- 0.17 &  91.87 +/- 0.57 &  95.03 +/- 0.58 &  96.80 +/- 0.08 &  96.57 +/- 0.12 \\
\citet{kim2021embedding} (SS) &  92.57 +/- 0.62 &  93.47 +/- 0.96 &  94.17 +/- 0.88 &  96.40 +/- 0.08 &  96.67 +/- 0.26 &  96.17 +/- 0.18 \\
SimCLR (SS) &  91.67 +/- 0.56 &  92.73 +/- 0.36 &  93.23 +/- 0.34 &  94.60 +/- 0.12 &  93.93 +/- 0.30 &  94.73 +/- 0.03 \\
SleepPriorCL (SS) &  92.00 +/- 0.64 &  94.20 +/- 0.50 &  95.27 +/- 0.73 &  96.43 +/- 0.42 &  96.93 +/- 0.28 &  \textbf{97.33 +/- 0.35} \\
\bottomrule
\end{tabular}
\caption{Tabular data for the semi-supervised comparison shown in Fig.\ \ref{fig:label_ratios_linear_new} for KNN ($k=1$) classification accuracies on the HAR dataset.}
\label{tab:mean_var_semi_har_knn}
}
\end{table}

\begin{table}[h]
\centering
\resizebox{0.99\textwidth}{!}{
\begin{tabular}{lllllll}
\toprule
{Labeled Data} &              5\% &             10\% &             20\% &             50\% &             70\% &            100\% \\
\midrule
ExpCLR (SS) &  \textbf{81.47 +/- 0.18} &  \textbf{81.33 +/- 0.05} &  \textbf{82.27 +/- 0.23} &  \textbf{82.50 +/- 0.17} &  \textbf{83.00 +/- 0.19} &  \textbf{83.00 +/- 0.25} \\
ExpCLR (S) &  73.30 +/- 0.17 &  76.93 +/- 0.44 &  77.50 +/- 0.63 &  79.93 +/- 0.12 &  81.47 +/- 0.40 &  82.30 +/- 0.05 \\
Cross-Entropy (S) &  69.80 +/- 0.54 &  72.77 +/- 0.67 &  75.47 +/- 0.20 &  79.17 +/- 0.26 &  80.17 +/- 0.10 &  80.80 +/- 0.22 \\
\citet{kim2021embedding} (SS) &  78.13 +/- 0.49 &  79.73 +/- 0.10 &  80.20 +/- 0.14 &  81.00 +/- 0.21 &  81.60 +/- 0.12 &  81.10 +/- 0.34 \\
SimCLR (SS) &  48.23 +/- 0.47 &  58.00 +/- 0.17 &  67.67 +/- 0.82 &  71.43 +/- 0.22 &  72.50 +/- 0.24 &  76.30 +/- 0.16 \\
SleepPriorCL (SS) &  77.90 +/- 0.31 &  77.87 +/- 0.45 &  78.27 +/- 0.11 &  79.10 +/- 0.09 &  78.83 +/- 0.24 &  79.47 +/- 0.20 \\
\bottomrule
\end{tabular}
\caption{Tabular data for the semi-supervised comparison shown in Fig.\ \ref{fig:label_ratios_linear_new} for KNN ($k=1$) classification accuracies on the SleepEDF dataset.}
\label{tab:mean_var_semi_sleep_knn}
}
\end{table}

\begin{table}[h]
\centering
\resizebox{0.99\textwidth}{!}{
\begin{tabular}{lllllll}
\toprule
{Labeled Data} &              5\% &             10\% &             20\% &             50\% &             70\% &            100\% \\
\midrule
ExpCLR (SS) &  94.07 +/- 0.38 &  \textbf{95.53 +/- 1.00} &  \textbf{95.20 +/- 1.56} &  \textbf{96.33 +/- 0.79} &  \textbf{97.13 +/- 0.43} &  97.07 +/- 0.57 \\
ExpCLR (S) &  91.97 +/- 1.81 &  92.27 +/- 1.70 &  89.20 +/- 1.10 &  92.63 +/- 1.17 &  97.10 +/- 0.50 &  \textbf{97.67 +/- 0.10} \\
Cross-Entropy (S) &  89.70 +/- 1.91 &  90.50 +/- 1.41 &  93.83 +/- 0.47 &  93.67 +/- 0.68 &  92.63 +/- 2.99 &  96.97 +/- 0.14 \\
\citet{kim2021embedding} (SS) &  78.17 +/- 4.75 &  78.67 +/- 6.27 &  78.40 +/- 3.58 &  89.10 +/- 3.42 &  90.33 +/- 3.12 &  90.97 +/- 2.26 \\
SimCLR (SS) &  91.43 +/- 2.35 &  93.90 +/- 1.53 &  93.80 +/- 1.03 &  93.90 +/- 1.26 &  94.70 +/- 1.21 &  95.90 +/- 1.35 \\
SleepPriorCL (SS) &  \textbf{96.10 +/- 0.58} &  94.10 +/- 1.67 &  94.63 +/- 0.66 &  95.67 +/- 1.05 &  94.30 +/- 1.18 &  95.33 +/- 1.16 \\
\bottomrule
\end{tabular}

\caption{Tabular data for the semi-supervised comparison shown in Fig.\ \ref{fig:label_ratios_linear_new} for KNN ($k=1$) classification accuracies on the Waveform dataset.}
\label{tab:mean_var_semi_waveform_knn}
}
\end{table}

\section{Experimental Details}\label{sec:detail_exp}
\subsection{Datasets}\label{sec:detail_datasets}
In the following we list the sources, further information, and implementation details of the datasets and expert features we used in our work.

\textbf{HAR:} The HAR dataset aims to classify six different activity states of humans, namely: walking, walking upstairs, walking downstairs, sitting, standing, laying. They collected the data using a mounted Samsung Galaxy S2 device where a triaxial acceleration and gyroscope sensor is installed. We downloaded the dataset from the UCI Machine Learning Repository (\emph{https://archive.ics.uci.edu/ml/datasets/human+activity+recognition+using+smartphones}) and preprocessed the data as \cite{eldele2021time} did in the corresponding repository (\emph{https://github.com/emadeldeen24/TS-TCC}).

\textbf{SleepEDF:} For sleep stage classification they differ five different classes: Wake (W), Non-rapid eye movement (N1,
N2, N3) and Rapid Eye Movement (REM). We downloaded the dataset from the PyhsioNet database (\emph{https://physionet.org/content/sleep-edf/1.0.0}) and  loade/preprocessed the data like \cite{eldele2021time} did in the repository (\emph{https://github.com/emadeldeen24/TS-TCC}).

\textbf{Waveform:} This dataset distinguishes four different classes: Artial Fibrillation (AFIB), atrial flutter (AFL), AV junctional rhythm (J), and all other rhythms (N). We downloaded the data from the PyhsioNet database (\emph{https://physionet.org/content/afdb/1.0.0}) and preprocessed it like \cite{tonekaboni2020unsupervised} did in the corresponding repository (\emph{https://github.com/sanatonek/TNC\_representation\_learning}).

\subsection{Expert Features:}\label{sec:detail_expert}
\textbf{HAR:} To obtain expert features, \cite{cruciani2019public} first filtered the raw time-series signals in order to reduce noise. Subsequently, for selected signals they compute signal magnitudes and applied a Fast Fourier Transform. In order to equip all the resulting signals with expert features, they compute scalar attributes like maximimal value, minimal value, means, energy or standard deviation for each signal. Finally they end up with a 561-dimensional expert feature vector for each sample.

\textbf{SleepEDF:} Following \cite{HUANG2020105253} we select their proposed methods to calculate expert features for sleep stage classification from ECG signals. They identify 30 suitable features from time and frequency domain for this classification task which we implemented in our repository. We couldn't find implementation details of the fractal dimension and left this feature concluding with a 29 dimensional expert feature vector.

\textbf{Waveform:} For this dataset \cite{goodfellow2017classification} found representative features for classification of ECG signals. They distinguish three feature types: Full waveform features which are extracted from the wavelet transformation, template features which identify medical properties of the ECG signal and mainly separate between normal rythm and artial fibrillation, and lastly RRI features which identify properties of important signal peaks. For our model we used the full waveform features and the RRI features as the template features are not directly implemented in \cite{goodfellow2017classification} repository (\emph{https://github.com/Seb-Good/ecg-features/blob/f9a4c986f8e460a081c71b8e2c7e3ddb26eabae8}). Finally we get a 176 dimensional feature vector for each sample.

\subsection{Implementation Details}\label{sec:imp_details_app}

\textbf{Model:} We start from TCN implementation \cite{bai2018empirical} and fit the structure in order to remove time causality and same sequence length output, as we generally want a lower dimensional embedding. We apply a constant channel- and kernel size over all convolutional layers. We only increase the stride downsampling for longer sample lengths as they occur in SleepEDF and Waveform dataset.

\textbf{Training:} For both, ExpCLR and the competing methods we choose individual learning rates for unsupervised and supervised training for each dataset optimzed in a range of $lr \in$ \text{\{}$5\mathrm{e}{-5}$, $1\mathrm{e}{-4}$, $5\mathrm{e}{-4}$, $1\mathrm{e}{-3}$, $3\mathrm{e}{-3}$, $5\mathrm{e}{-3}$, $7\mathrm{e}{-3}$, $1\mathrm{e}{-2}$\text{\}} each. As performance indicator we choose linear accuracy as we did in Sec. \ref{sec:imp_details}. The final learning rates for each algorithm for supervised (S) and unsupervised (U) training modes and each dataset are shown in Tab. \ref{tab:learning_rates}. Further we investigate different values for the parameters $\tau$ and $\Delta$ from Eq. \ref{eq:expclr_loss}. To avoid overfitting ExpCLR w.r.t our competing methods we set $\tau$ and $\Delta$ to be the same for all datasets. Regarding training stability and accuracy we identify $\tau=1$ and $\Delta=1$ as a good choice.

\begin{table}[!ht]
\centering
\begin{tabular}{lllllll}
\toprule
{Dataset and $lr$} & HAR (U) & HAR (S) & SleepEDF (U) & SleepEDF (S) & Waveform (U) & Waveform (S)  \\
\midrule
ExpCLR &  $3\mathrm{e}{-3}$ & $1\mathrm{e}{-3}$ & $5\mathrm{e}{-3}$ & $1\mathrm{e}{-2}$ & $5\mathrm{e}{-3}$ & $5\mathrm{e}{-3}$ \\
\citet{kim2021embedding} & $3\mathrm{e}{-3}$ & $1\mathrm{e}{-3}$ & $5\mathrm{e}{-3}$ & $1\mathrm{e}{-2}$ & $7\mathrm{e}{-3}$ & $5\mathrm{e}{-3}$\\
SimCLR &  $3\mathrm{e}{-3}$ & $5\mathrm{e}{-4}$ & $5\mathrm{e}{-3}$ & $1\mathrm{e}{-5}$ & $7\mathrm{e}{-3}$ & $5\mathrm{e}{-4}$\\
SleepPriorCL &  $1\mathrm{e}{-3}$ & $1\mathrm{e}{-3}$ & $7\mathrm{e}{-3}$ & $1\mathrm{e}{-3}$ & $1\mathrm{e}{-2}$ & $1\mathrm{e}{-3}$\\
Cross-Entropy &  - & $5\mathrm{e}{-4}$ & - & $1\mathrm{e}{-3}$ & - & $1\mathrm{e}{-3}$\\
\bottomrule
\end{tabular}
\caption{Learning rates for each algorithm and dataset in unsupervised (U) and supervised (S) training mode, which we identified as best via a grid search. Note that some approaches reduce to a different loss during supervised training mode (see Sec. \ref{sec:competing_methods})}
\label{tab:learning_rates}
\end{table}

\subsection{Competing Methods:}\label{sec:comp_details}

For all methods except TS-TCC, we implemented the methods in our repository as explained in Sec.\ \ref{sec:competing_methods}. For the comparison we only substitute the loss function used during optimization and keep the architecture and evaluation methods the same for all competing methods. As explained in the previous section, we select the learning rate based on a grid-search and choose the best learning rate individually for each combination of algorithm and dataset. Apart from this, we did not conduct an extensive search of all the respective hyperparameters but rather chose natural ones or used the parameters as described in the original works. This also applies to ExpCLR as can be seen in Fig.\ \ref{fig:ablation_loss_tau}, where other $\tau < 1.0$ could improve the performance slightly. For the data transformations we tried out different combination and found dropout and scaling to work well for the respective datasets. These data transformations where then used for SimCLR.

For TS-TCC we made use of the already available repository (\emph{https://github.com/emadeldeen24/TS-TCC}). We adapted the repository to include the exact same datasets and seeds we used and added our Waveform dataset to the repository. Further, we tried replacing the encoder used by TS-TCC with our encoder architecture ,but found the performance to drop significantly. Therefore, we chose to keep the original encoder architecture.

\section{Proofs}\label{sec:proofs}
\subsection{Proof of Proposition \ref{prop:prop1}}\label{sec:proof_prop_1}

Let us first consider the case $\text{dim}(\text{ker}(M_\phi))=0$. Further, we assume the input points to be generic and thus also the embedding vectors $E_i := E_\theta(x_i)$ can be considered generic. Since our goal is to show that the condition $\mathcal{L}_{mse}(M_\phi\circ E_\theta(X),F)=0$ is inufficient to derive any bound on $l_-$ and $l_+$, we assume that there exist fixed $B_-,B_+\in \mathbb{R_+}$ such that $B_- \leq l_-$ and $B_+ \geq l_+$ for all encoders $E_\theta$ and linear maps $M_\phi$ which satisfy $\mathcal{L}_{mse}(M_\phi\circ E_\theta(X),F)=0$. Next we set $\phi$ and $\theta$ to a value for which $\mathcal{L}_{mse}(M_\phi\circ E_\theta(X),F)=0$, i.e.\ $M_\phi(E_\theta(x_i))=f_i$ reproduces the expert features $f_i$ on each datapoint $x_i$. Thus, by Def.\ \ref{definition1} the best possible values for $l_-$ and $l_+$ are
\begin{equation}
    l_- = \min_{i,j} \frac{\| M_\phi ( E_i-E_j)\|_2}{\|E_i-E_j\|_2} \quad \text{and} \quad l_+ = \max_{i,j} \frac{\| M_\phi( E_i-E_j)\|_2}{\|E_i-E_j\|_2},
\end{equation}
where $i,j$ run over all pairs in $\{1,\dots,N\}$ with $E_i\neq E_j$. These values $l_-$ and $l_+$ are strictly positive due to $\text{dim}(\text{ker}(M_\phi))=0$. 
Next, we use that the encoder is a neural network (NN) with $L$ layers, so that the weights $\theta$ can be grouped as $\theta = (\theta_1,\dots,\theta_L)$, where $\theta_i$ are the weights of the $i$-th layer. Then define $\widetilde{\theta}:=(\theta_1,\dots,\theta_{L-1},\frac{1}{c}\theta_L)$ and $\widetilde{\phi}:=c\phi$ for a constant $c\in{\mathbb R}_+$, where $\phi$ directly parameterizes the matrix $M_\phi$. Since $E_\theta\circ M_\phi = E_{\widetilde{\theta}}\circ M_{\widetilde{\phi}}$, it also holds that ${\mathcal L}_{mse}(M_{\widetilde{\phi}}\circ E_{\tilde{\theta}}(X),F)=0$ attains the global loss minimum, for any $c\in{\mathbb R}^+$. 
While this rescaling does not change the NN output as the two factors cancel, it changes $l_-$ to $\widetilde{l}_-= c l_-$  and $l_+$ to $ \widetilde{l}_+ = c l_+$. Thus, by choosing $c\in{\mathbb R}_+$ appropriately, one can attain any positive values for $l_-$ or $l_+$ while still assuring that $\mathcal{L}_{mse}(M_\phi\circ E_\theta(X),F)=0$. Therefore as claimed, it is not possible to provide any guarantees or bounds $B_-,B_-$ on $l_-,l_+$ solely from the condition $\mathcal{L}_{mse}(M_\phi\circ E_\theta(X),F)=0$.

Next, let us consider the case $\text{dim}(\text{ker}(M_\phi)) > 0$. This can be either due to the specific choice of $\phi$ or hold generally when $e > d$ since one has $\text{dim}(\text{ker}(M_\phi))\geq e-d > 0$. In this case the embedding is not unique for a given expert feature vector $f_i$ since for any $\widetilde{E} \in \text{ker}(M_\phi)$ we have $M_\phi(E_i+\widetilde{E})= M_\phi(E_i) = f_i$. Now let us assume there exists a constant $l_->0$ such that for all embeddings which satisfy $M_\phi(E_i) = f_i\ \forall i$ (which is equivalent to ${\mathcal L}_{mse}(M_\phi\circ E_\theta(X),F)=0$) it holds that $l_- \|E_i -E_j\|_2 \leq \| f_i-f_j\|_2\ \forall i,j$. Next, fix such an embedding $E_\theta$ and consider a fixed pair $k,l \in\{1,\dots, N\}$ with $f_k\neq f_l$. 
Then choose $\widetilde{E}\in \text{ker}(M_\phi)$ such that $\|\widetilde{E}\|_2 \geq \frac{2}{l_-}\| f_k-f_l\|_2+\|E_k -E_l\|_2$. Then a neural network embedding $E'$ can be constructed that satisfies $E'(x_k)=E_k+\widetilde{E}$ and $E'(x_i)=E_i\ \forall i\neq k$. Note that $M_\phi(E'(x_i))=M_\phi(E_\theta(x_i))\ \forall i\in\{1,\ldots,N\}$, so that also ${\mathcal L}_{mse}(M_\phi\circ E'(X),F)=0$. 
However, $E'$ does \emph{not} satisfy the aforementioned bilipschitz inequality for $l_-$ since by the triangle inequality it holds
\begin{equation*}
    l_-\|E'(x_k)-E'(x_l)\|_2 = l_- \|E_k +\tilde{E} -E_l\|_2 \geq l_-\left(\|\tilde{E}\|-\|E_k-E_l\|\right)\geq 2\| f_k-f_l\|_2>\|f_k-f_l\|_2,
\end{equation*}
therefore violating the inequality. Thus, solely from $\mathcal{L}_{mse}(M_\phi\circ E_\theta(X),F)=0$ a non-zero value for $l_-$ cannot be concluded, and thus it is not guaranteed that $E_\theta$ is a bilipschitz embedding.
\QED

\subsection{Proof of Proposition \ref{proposition_2}}\label{sec:proof_prop_2}
From $\mathcal{L}_{quad}(E(X),F)=0$ it directly follows that $\forall i,j \in\{1,\dots, N\}$ we have $\left((1-s_{ij})\Delta -D_{ij}\right)=0$ and thus
\begin{equation*}
    \frac{\Delta}{\max_{k,l}\|f_k-f_l\|_2} \|f_i-f_j\|_2 = \|E_i-E_j\|_2.
\end{equation*}
Therefore, one can set $l_- = l_+ = \frac{\max_{k,l}\|f_k-f_l\|_2}{\Delta}$.
\QED

While the similarity measure used in Proposition \ref{proposition_2} provides very good bounds on $l_-$ and $l_+$ it is easy to see that other similarity measures like the one introduced in Sec.\ \ref{sec:practical_aspects} also provide good bounds for $l_+$ and $l_-$.

\subsection{Proof of Proposition \ref{proposition_3}}\label{sec:proof_prop_3}
We start be proving (a) and then move on to (b).  (a) can be easily shown by making use of the softmax limit
\begin{equation*}
    \tau \ln{\sum_i \exp{\left(\frac{a_i}{\tau}\right)}} \to \max_i a_i\qquad\text{as}~~\tau\to0,
\end{equation*}
for $a_i \in \mathbb{R}$. Using this one can directly see that
\begin{equation*}
    \tau \log\left[\frac{1}{N^2}\sum_{i,j}\exp(L_{ij}/\tau)\right] \to \max_{i,j}(L_{ij}).
\end{equation*}
For (b) and the limit $\tau\to\infty$, we start by using a series-expansion of $L_{ExpCLR}^\tau(E(X),F)$ which gives
\begin{equation*}
    L_{ExpCLR}^\tau(E(X),F) =  \tau\log\left[\frac{1}{N^2}\sum_{i,j}\left(1+\frac{L_{ij}}{\tau}+O\left(\frac{1}{\tau^2}\right)\right)\right]=\frac{1}{N^2}\sum_{i,j}L_{ij} + \mathcal{O}\left(\frac{1}{\tau}\right).
\end{equation*}
And thus for $\tau \to \infty$ we get   $L_{ExpCLR}^\tau(E(X),F) \to  \frac{1}{N^2}\sum_{i,j}L_{ij}$.
\QED

\section{Statistical Bounds}\label{sec:statistical_bounds}
The loss $ L_{ExpCLR}^\tau(E(X),F) $ proposed by our work (see Eq.\ \ref{eq:expclr_loss}) mostly focuses on the pairs with largest loss value due to the hard negative mining (when $\tau$ is small). If we define the pair-Lipschitz constant to be $Z_{ij}:= \frac{\|f_i-f_j\|_2}{ \| E(x_i)-E(x_j)\|_2}$, the pairs with largest loss tend to be the ones whose pair-Lipschitz constant is furthest away from $ \frac{\max_{k,l}\|f_k-f_l\|_2}{\Delta}$. Defining $l_{min} = \min_{k,l}Z_{kl}$ and $l_{max} = \max_{k,l}Z_{kl}$, we have $Z_{ij} \in [l_{min},l_{max}]\ \forall i,j$; here $i,j$ run over all indices in the respective training set. As explained in the methods section (Prop.\ \ref{proposition_2}), during training our loss tries to push $l_{min}$ and $l_{max}$ close to $\max_{kl}\|f_k-f_l\|/\Delta$.

While so far this only guarantees the pair-Lipschitz constants $Z_{ij}$ from the training set to lie in the interval $[l_{min},l_{max}]$, the question arises, whether we can guarantee that the $Z_{ij}$ are also inside a certain interval for new unseen samples $x_i,x_j$.

We therefore split our dataset into two sets $D_{train}$ and  $D_{val}$ of pairs of datapoints, where $D_{train}$ consists of $N_{train}$ pairs of i.i.d.\ samples and $D_{val}$ of $N_{val}$ pairs of i.i.d.\ samples. The pair-Lipschitz constant is thus a random variable, determined by sampling two i.i.d.\ samples $x_i, x_j$ from the underlying data distribution ${\mathbb P}$ and then evaluating $Z_{ij}:= \frac{\|f_i-f_j\|_2}{ \| E(x_i)-E(x_j)\|_2}$, where $f_i,f_j$ are the corresponding expert features of the input samples and $E(x_i),E(x_j)$ denotes the representations. The encoder $E$ is updated by minimizing a loss function. We now present two approaches to obtain statistical bounds on the pair-Lipschitz constant for new unseen pairs of samples.

\subsection{First Approach: Interval boundaries from validation set}\label{PAC-first}

1) Train an encoder $E$ by minimizing the loss over $D_{train}$.

2) Calculate $l_{min}^{val}$ and $l_{max}^{val}$ as the minimum and maximum over the $Z_{ij}$ on the validation set $D_{val}$.

3) Then we get via a PAC-bound that with prob. $(1-\delta)$ w.r.t. repeated sampling of the validation set we have 
\begin{equation}
P(Z_{test} \notin [l_{min}^{val},l_{max}^{val}]) \leq \sqrt{\frac{8\ln\left[2N_{val}(2N_{val}-1)\frac{4}{\delta}\right]}{N_{val}}}, 
\end{equation}
where $Z_{test}:= \frac{\|f-f'\|_2}{ \| E(x)-E(x')\|_2}$ is the pair-Lipschitz constant obtained on two unseen (test) i.i.d.\ samples $x$ and $x'$.

\subsection{Second Approach: Interval boundaries from training set}\label{PAC-second}

1) Train an encoder $E$ by minimizing the loss over $D_{train}$ and calculate $l_{min}^{train}$ and $l_{max}^{train}$ as the minimum and maximum over the $Z_{ij}$ on the training set $D_{train}$.

2) Calculate $P_{val}:= \frac{1}{N_{val}}\sum_{(i,j)\in D_{val}}1_{Z_{ij} \notin [l_{min}^{train},l_{max}^{train}]}$ on $D_{val}$.

3) Then again via a PAC-bound we get that with prob. $(1-\delta)$ w.r.t. repeated sampling of the validation set  
\begin{equation}
P(Z_{test}\notin [l_{min}^{train},l_{max}^{train}]) \leq P_{val} + \sqrt{\frac{\ln\left[\frac{2}{\delta}\right]}{N_{val}}},
\end{equation}
where again $Z_{test}:= \frac{\|f-f'\|_2}{ \| E(x)-E(x')\|_2}$ is the pair-Lipschitz constant obtained on two unseen (test) i.i.d.\ samples $x$ and $x'$. 
\begin{figure}[h]
\center
\includegraphics[scale=0.5]{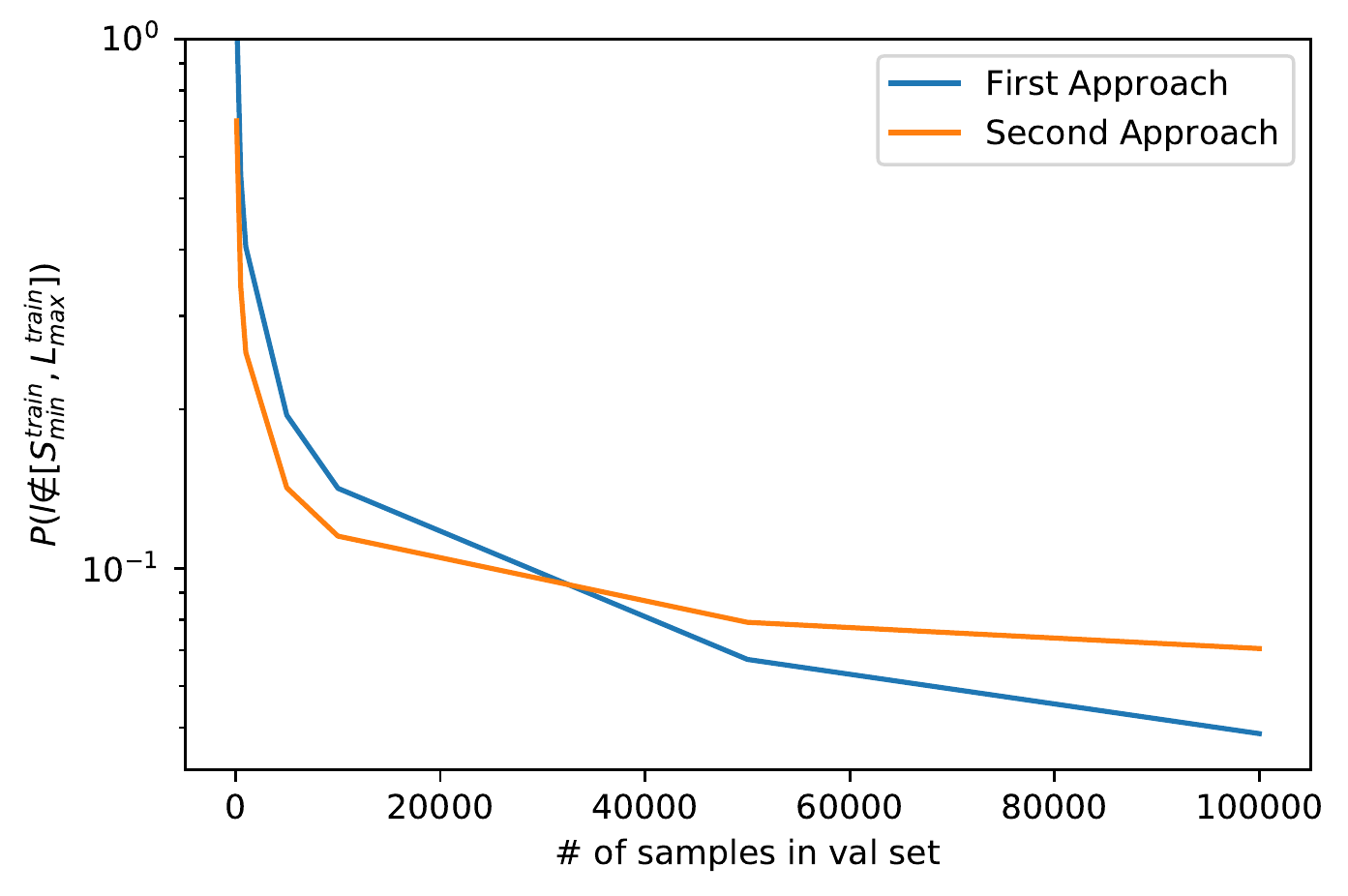}
\caption{\textbf{Empirical comparison of the two bounds from Apps.\ \ref{PAC-first} resp.\ \ref{PAC-second}:}
The figure shows the upper bounds on the probability that the pair-Lipschitz constant lies outside $[l_{max},l_{min}]$ for a randomly drawn test pair, over the number of pairs in the validation set $N_{val}$. The empirical evaluation shows that for a large number of validation pairs the first approach leads to better guarantees, while for smaller number of pairs the second approach beats the first approach.
}
\label{fig:emperical_validatrion}
\end{figure}
\subsection{Empirical Evaluation of the Bounds}
We now empirically evaluate which of the two previous bounds gives better guarantees. Therefore, we assume to have a discrepancy between training and validation bounds which leads to $5\%$ of the validation pairs to have sampled $Z_{ij}$ outside of the training bounds $[l_{min}^{train},l_{max}^{train}]$ and also that we have some overfitting which leads to a smaller interval on the training set. Thus, for illustration purposes, we assume $P_{val} \approx  0.05$, $l_{min}^{val} \leq l_{min}^{train}$ and  $l_{max}^{val} \geq l_{max}^{train}$ .

We evaluate both bounds for different numbers of pairs in the validation set $N_{val}$ and calculate our bounds on the probability of a sampled $Z_{test}$, generated by a pair of i.i.d.\ samples, to lie outside the window $[l_{max},l_{min}]$. (Here we consider the case where we are mostly interested in the probability bounds, not so much in the exact interval boundaries.) The comparison over the number of validation samples is shown in Fig.\ \ref{fig:emperical_validatrion}.

If one only requires an upper \emph{or} a lower bound on the pair-Lipschitz constant, the bound can be improved since as growth function reduces to $G(n)= n$. This then improves the bound from App.\ \ref{PAC-first} slightly: In this case with probability at least $(1-\delta)$ w.r.t.\ repeated sampling of the validation set we have 
\begin{equation*}
P(Z_{test} \geq l_{max}^{val}) \leq \sqrt{\frac{8\ln\left[\frac{8 N_{val}}{\delta}\right]}{N_{val}}}.
\end{equation*}

\section{Comparison of Loss Function Variants and their Derivatives}\label{sec:loss_functions_cont_exp}

\begin{figure*}[h]
\center
\includegraphics[scale=0.65]{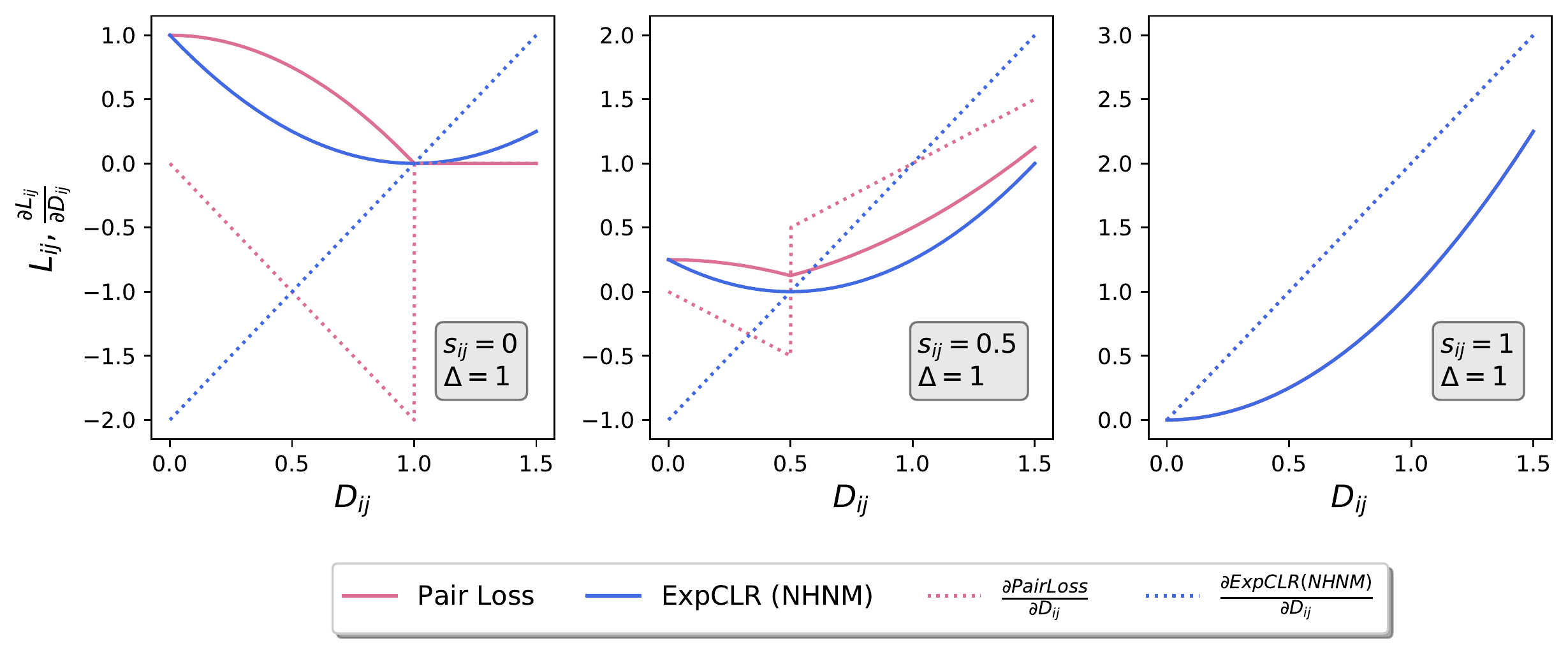}
\caption{\textbf{Comparison of ExpCLR (NHNM) and the Pair-Loss with their associated Derivatives:} This figure shows the Loss function of ExpCLR (NHNM=no hard-negative mining) from Eq.\ \ref{eq:quadratic_con_loss} and the pair-loss from Eq.\ \ref{eq:pair_loss} and their derivatives. On the y-axis one can see values for the loss function resp. derivative of the loss function and on the x-axis certain values for the distance metric $D_{ij}$ are given. From the left to the right panel different scalar values of the similarity measure $s_{ij}$ are shown. Note that in right plot both losses reduce to the same as $s_{ij} = 1$.
}
\label{fig:derivatives_loss}
\end{figure*}

Below one can find the equations for ExpCLR (NHNM) and the pair loss with their respective derivatives. Fig. \ref{fig:derivatives_loss} additionally visualizes them over a range of values for the distance metric $D_{ij}$. One can see that ExpCLR has the same minimum as the pair loss, but possesses continuous derivatives w.r.t. $D_{ij}$.

\begin{equation*}
    {\mathcal L}_{ij, ExpCLR(NHNM)} = 
    \big((1-s_{ij})\Delta -D_{ij} \big)^2
\end{equation*}

\begin{equation*}
    \frac{\partial {\mathcal L}_{ij, ExpCLR(NHNM)}}{\partial D_{ij}} = 
    -2((1-s_{ij})\Delta-D_{ij})
\end{equation*}

\begin{equation*}
    {\mathcal L}_{ij, pair} = s_{ij}D_{ij}^2+\max{(0,(1-s_{ij})^2\Delta^2-D_{ij}^2)}
\end{equation*}

\begin{equation*}
    \frac{\partial {\mathcal L}_{ij, pair}}{\partial D_{ij}} = 
    \begin{cases}
        2s_{ij}D_{ij} & D_{ij} \geq \Delta(1-s_{ij}) \\
        -2D_{ij}(1-s_{ij}) &  else
    \end{cases}
\end{equation*}

\section{Gradient-Level Proof of Hard-Negative Mining}\label{sec:gradient_level_proof_hnm}
In this section we demonstrate on gradient-level how our hard-negative mining scheme introduced in Sec.\ \ref{sec:hard_negative_mining} functions. This complements the intuitions provided by Prop.\ \ref{proposition_3}. For this, we take the gradient of Eq.\ \ref{eq:expclr_loss}:
\begin{align*}\label{eq:expclr_loss_original}
     \frac{\partial }{\partial L_{nm}}{\mathcal L}_{ExpCLR}^\tau(E(X),F) &= \frac{\partial }{\partial L_{nm}} \tau \log\left[\sum_{i,j=1}^N\frac{\exp\left(\frac{L_{ij}}{\tau}\right)}{N^2}\right] \\
     &=  \exp\left(\frac{L_{nm}}{\tau}\right)\left[\sum_{i,j=1}^N\exp\left(\frac{L_{ij}}{\tau}\right)\right]^{-1}.
\end{align*}
Therfore, the gradient is directly proportional to $\exp\left(\frac{L_{nm}}{\tau}\right)$, adding an exponential (softmax) scaling and thus increasing the gradient contributions of those pairs $(n,m)$ with larger loss components $L_{nm}$.

\end{document}